\pdfoutput=1

\documentclass[11pt]{article}

\usepackage[preprint]{acl}

\usepackage{times}
\usepackage{latexsym}

\usepackage[T1]{fontenc}

\usepackage[utf8]{inputenc}

\usepackage{microtype}

\usepackage{inconsolata}

\usepackage{graphicx}
\usepackage{arydshln} 
\usepackage{booktabs}
\usepackage{multirow}
\usepackage{tabularx}   
\usepackage[table]{xcolor}
\usepackage{adjustbox}
\usepackage{enumitem}
\usepackage{amsmath}
\usepackage{amssymb}
\usepackage{pifont}
\usepackage{array}
\usepackage{geometry}
\usepackage{makecell}
\usepackage{footmisc}
\usepackage{CJKutf8}
\usepackage{hyperref}
\usepackage{subcaption}
\usepackage{url}

\title{MMDS-Bench: Benchmarking Multimodal Large Language Models on Dynamic Stance in Social Media Interactions}

\author{
  \textbf{Yuzhe Ding\textsuperscript{1}\thanks{Equal contribution.}},
  \textbf{Kang He\textsuperscript{1,2}\footnotemark[1]},
  \textbf{Li Zheng\textsuperscript{1}},
  \textbf{Shengwu Zheng\textsuperscript{1}},
  \textbf{Teng Shi\textsuperscript{1}},
\\
  \textbf{Fei Li\textsuperscript{1}},
  \textbf{Chong Teng\textsuperscript{1}},
  \textbf{Donghong Ji\textsuperscript{1}\thanks{Corresponding author.}}
\\
\\
  \textsuperscript{1}School of Cyber Science and Engineering, Wuhan University
\\
  \textsuperscript{2}Shanghai Innovation Institute
\\
  \texttt{\{yuzheding, hekang0225, dhji\}@whu.edu.cn}
}

\begin{document}
\maketitle

\begin{abstract}
Dynamic stance classification models how a reply responds to its direct parent message, rather than how a post relates to a fixed topic. Existing work has mainly studied this problem in text-only settings, while social media interactions increasingly rely on images, screenshots, memes, reaction images, and cross-modal references. We introduce MMDS-Bench\footnote{https://github.com/whu-yzding/MMDS-Bench}, a diagnostic benchmark for multimodal dynamic stance classification in social media parent--reply interactions. MMDS-Bench contains 3,482 multimodal instances annotated with a seven-label dynamic stance taxonomy, together with an 800-instance diagnostic subset that requires structured reasoning over parent understanding, reply understanding, and stance-relation inference. We further annotate each instance with five challenge factors covering multimodal fusion, parent framing, non-literal expression, interaction reasoning, and label-boundary ambiguity. We evaluate 12 closed-source and open-source multimodal large language models and propose a reference-grounded LLM-judge protocol for assessing reasoning quality. Results show that current MLLMs still struggle with multimodal dynamic stance understanding, especially in cases that require relational inference beyond separate parent and reply comprehension.
\end{abstract}

\section{Introduction}
Stance detection aims to identify opinions, attitudes, and disagreements in online discourse. Most prior work formulates the task as static or target-oriented stance detection, where a model predicts whether a post supports, opposes, or is neutral toward a predefined topic, target, or claim \citep{mohammad-etal-2016-semeval,sobhani-etal-2017-dataset,aldayel2021stance}. While useful for emotion recognition \citep{zhang-etal-2020-enhancing-cross,10.1145/3581783.3612053,zheng-etal-2026-dynamic,11300961}, rumor detection \citep{yu-etal-2020-coupled,yang-etal-2024-reinforcement}, sentiment analysis \citep{he2026pase,he-etal-2026-uncertainty,He_2026_CVPR}, and argument mining \citep{rajendran-etal-2018-something,sirrianni-etal-2020-agreement,10.1145/3539618.3591917}, this formulation does not fully capture the interactional nature of online discussions. In real conversations, users respond to specific previous messages by agreeing, disagreeing, elaborating, questioning, or shifting away from them. Dynamic stance classification addresses this setting by labeling how a reply responds to its direct parent message, rather than how a single post relates to a fixed global topic.

However, existing dynamic stance research \citep{figueras-etal-2023-dynamic,niu-etal-2024-challenge,ding-etal-2025-zero} has mainly focused on textual interactions, whereas social media conversations are increasingly multimodal \citep{liang-etal-2024-multi,niu2024multimodal}. Both parent and reply messages may contain text, images, screenshots, memes, reaction images, or their combinations. In such cases, stance may be expressed through visual cues, cross-modal references, non-literal expression, or multimodal irony. For example, a meme may signal disagreement only when interpreted against the parent message, while a screenshot may provide evidence that elaborates on a prior claim. These examples require models to understand both messages and infer how textual and visual signals jointly shape their conversational relation. We refer to this setting as \textit{multimodal dynamic stance classification}.

This task differs from standard multimodal understanding benchmarks \citep{10.1007/978-3-031-72658-3_13,Yue_2024_CVPR,pmlr-v235-yu24o}. The key challenge is not only recognizing objects, reading text in images, or describing visual content, but determining how multimodal cues function within a parent--reply interaction. A model must recover the parent framing, interpret the reply's communicative intent, and infer whether the reply agrees, disagrees, elaborates, queries, remains neutral, is unrelated, or cannot be reliably classified. Although recent multimodal large language models (MLLMs) have shown strong performance on visual question answering, OCR, captioning, and general multimodal reasoning, existing benchmarks rarely test whether they can recover such local stance relations in multimodal social media interactions.

To fill this gap, we introduce \textbf{MMDS-Bench}, a diagnostic benchmark for multimodal dynamic stance classification. MMDS-Bench contains 3,482 multimodal parent--reply instances, where both sides of the interaction include visual content and may also include text. Each instance is annotated with a seven-label dynamic stance taxonomy adapted from textual dynamic stance classification to multimodal social media interactions. MMDS-Bench contains two evaluation tasks. \textbf{Task 1}, Multimodal Dynamic Stance Classification, evaluates final-label prediction on the full benchmark. \textbf{Task 2}, Diagnostic Multimodal Dynamic Stance Reasoning, evaluates structured reasoning on an 800-instance diagnostic subset, requiring models to produce parent understanding, reply understanding, stance reasoning, and a final stance label. This design allows us to diagnose whether errors arise from misunderstanding either message or from failing to infer their stance relation.

We further annotate all instances with five challenge factors: Multimodal Fusion, Parent Framing, Non-Literal Reply, Interaction Reasoning, and Label-Boundary Ambiguity. These annotations support fine-grained analysis of how different multimodal and interactional difficulties affect model performance. We evaluate 12 closed-source and open-source MLLMs using final-label metrics and a reference-grounded LLM-judge protocol for reasoning quality. Our results show that multimodal dynamic stance classification remains challenging even for strong MLLMs. Models often understand the parent and reply separately, but still fail to infer how the reply dynamically responds to the parent. Performance also degrades as more challenge factors co-occur, suggesting that multimodal and interactional difficulties compound each other.

Our contributions are threefold:
\begin{itemize}
    \item We introduce \textbf{MMDS-Bench}, a diagnostic benchmark for multimodal dynamic stance classification in social media parent--reply interactions, containing 3,482 annotated instances and an 800-instance diagnostic subset.
    \item We provide a fine-grained evaluation framework that assesses both final stance labels and structured reasoning, together with five challenge-factor annotations for multimodal and interactional difficulty analysis.
    \item We evaluate 12 closed-source and open-source MLLMs and introduce a reference-grounded LLM-judge protocol for assessing reasoning quality beyond final-label accuracy.
\end{itemize}

\section{Related Work}

\paragraph{Stance Detection.}
Stance detection aims to identify the stance expressed in a text toward a given target \citep{augenstein-etal-2016-stance,DING2027105103}. Existing target-oriented settings are commonly categorized into in-target, cross-target, and zero-shot stance detection. In-target methods assume the same targets during training and evaluation \citep{mohammad-etal-2016-semeval,li-caragea-2019-multi,li-caragea-2021-target}, while cross-target methods transfer knowledge from source targets to related unseen targets \citep{xu-etal-2018-cross,zhang-etal-2020-enhancing-cross,li-etal-2021-improving-stance}. Zero-shot stance detection further aims to generalize to unseen targets without target-specific training examples \citep{liu-etal-2021-enhancing,liang-etal-2022-jointcl,luo-etal-2022-exploiting}.
Recent studies have also explored the capabilities of large language models (LLMs) for stance detection \citep{HUANG2026132677}. LLMs enable stance prediction through prompting and in-context learning without task-specific training, and have been investigated in open-target and zero-shot settings \citep{akash-etal-2025-large,nguyen-kim-2025-external,li-etal-2025-mitigating-biases}. However, their predictions can still be affected by external information, sentiment--stance correlations, and target-related biases.

Despite these advances, target-oriented stance detection still primarily models the relation between a single post and an external target. This formulation is less suitable for local conversational interactions, where the key question is often how a reply responds to its direct parent message rather than whether an individual post supports or opposes a predefined topic. This motivates the dynamic stance formulation discussed next.

\begin{table*}[t]
\centering
\small
\begin{tabularx}{\linewidth}{p{0.32\linewidth}X}
\toprule
\textbf{Challenge Factor} & \textbf{Annotation Criterion} \\
\midrule
Multimodal Fusion &
The image itself is difficult to interpret, or the text--image relation is complex, making the overall meaning difficult to determine reliably. \\

Parent Framing &
The parent lacks an explicit proposition, target, or tone, making it difficult to determine which aspect of the parent the reply addresses. \\

Non-Literal Reply &
The stance is mainly expressed through sarcasm, memes, exaggeration, implication, analogy, or other non-literal forms. \\

Interaction Reasoning &
The stance relation between the reply and the parent requires complex internal or external context understanding, or multi-step relation mapping. \\

Label-Boundary Ambiguity &
The instance satisfies the criteria of multiple stance labels, making single-label classification unstable. \\
\bottomrule
\end{tabularx}
\caption{Challenge factors annotated in MMDS-Bench.}
\label{tab:challenge_factors}
\end{table*}

\paragraph{Dynamic Stance Detection.}
In real-world social media scenarios, users often express opinions through
conversational interactions rather than isolated posts. This has motivated
conversational stance detection, which aims to identify stance expressions within discussion threads. Early work such as the SRQ dataset studied stance in comment data, but mainly focused on single-turn replies and relatively shallow conversational structures \citep{villa2020stance}. More recently, MT-CSD expanded conversational stance detection to deeper multi-turn conversations across five targets \citep{niu-etal-2024-challenge}, MmMtCSD extended this line of work to multimodal conversation scenarios \citep{niu2024multimodal}, and MT2-CSD further scaled up multi-turn conversational stance detection with a substantially larger human-annotated dataset featuring deeper conversation threads and an additional post-as-target setting \citep{NIU2026109304}.

These studies show the importance of modeling stance in conversational context. However, many conversational stance datasets still retain a target-centered or claim-centered formulation, where the stance is defined with respect to an external topic, target, or claim. In contrast, dynamic stance classification defines stance as the local relation between a reply and its direct parent message, capturing how the reply agrees, disagrees, elaborates, queries, or otherwise responds to the parent \citep{figueras-etal-2023-dynamic}.
This interaction-centered formulation is closer to how stance is expressed in
threaded social media discussions. Nevertheless, existing dynamic stance work has mainly focused on text-only interactions, leaving multimodal parent--reply relations underexplored.

\paragraph{Multimodal Stance and Social Media Understanding.}
Multimodal stance detection extends stance analysis beyond text by jointly modeling textual and visual evidence in social media content. Early studies introduced multimodal stance resources and demonstrated the importance of cross-modal interactions for stance inference \citep{weinzierl-harabagiu-2023-identification,liang-etal-2024-multi}. Subsequent work expanded multimodal stance detection to conversational and video-based scenarios \citep{niu2024multimodal,wang-etal-2024-multiclimate}, while recent methods have explored target-aware cross-modal alignment, explicit multimodal reasoning, and retrieval- or agent-enhanced inference \citep{zhang-etal-2025-mad,wang-etal-2026-mind,lu-etal-2026-mm}. Related multimodal social media studies further highlight the importance of jointly interpreting visually grounded cues such as memes, screenshots, and other image--text combinations \citep{zheng2025multi}.

Despite this progress, most multimodal stance benchmarks remain target-oriented, predicting the stance of multimodal content toward an external target, topic, or claim. Even multimodal conversational stance detection primarily models how individual utterances relate to predefined targets rather than how a reply responds to its direct parent \citep{niu2024multimodal}. In contrast, multimodal dynamic stance requires modeling the local relation between two multimodal messages, where the interpretation of a reply may change with the parent message it responds to.

\begin{figure*}[t]
\centering
\includegraphics[width=1\linewidth]{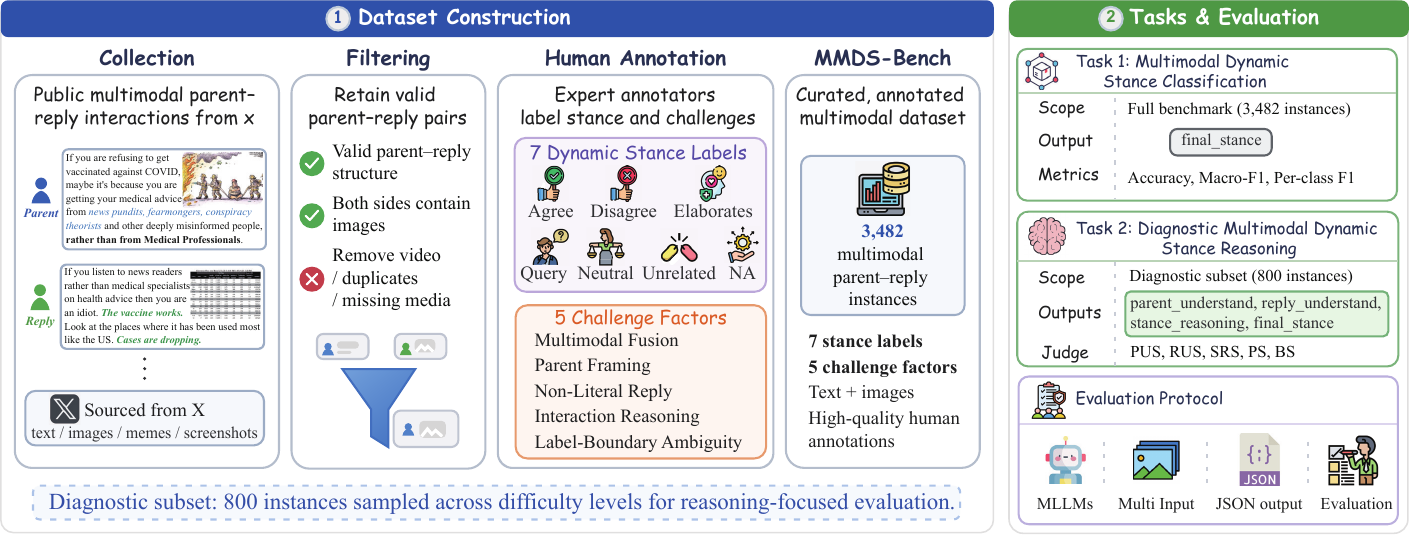}
\caption{Overview of the dataset construction, tasks, and evaluation in MMDS-Bench.}
\label{fig:mmds_bench}
\end{figure*}

\section{MMDS-Bench}
Figure~\ref{fig:mmds_bench} illustrates the overall workflow of MMDS-Bench, from multimodal parent--reply data collection to annotation, evaluation, and diagnostic analysis.

\subsection{Task Definition}
MMDS-Bench evaluates multimodal dynamic stance classification in social media parent--reply interactions. Given a parent message and its direct reply, the task is to determine how the reply responds to the parent, using textual, visual, and cross-modal cues. We adopt the seven-label taxonomy proposed by \citet{figueras-etal-2023-dynamic}: Agree, Disagree, Elaborates, Query, Neutral, Unrelated, and NA. The taxonomy is extended from text-only conversations to multimodal interactions involving screenshots, memes, reaction images, and visual evidence. When multiple stance signals appear, the label reflects the most salient reply-to-parent relation.

MMDS-Bench includes two tasks. \textbf{Task 1}, Multimodal Dynamic Stance Classification, evaluates final-label prediction on the full benchmark. \textbf{Task 2}, Diagnostic Multimodal Dynamic Stance Reasoning, requires models to additionally output \texttt{parent\_understanding}, \texttt{reply\_understanding}, and \texttt{stance\_reasoning}. This structured format helps diagnose whether errors come from parent understanding, reply understanding, or stance-relation inference.

\subsection{Dataset Construction}
\paragraph{Data Collection.}
We collect MMDS-Bench from public parent--reply interactions on X\footnote{https://x.com/}. To cover diverse social discussions, we define a set of domain-specific keywords across topics such as politics, public health, climate, immigration, gender, economy, and social events, with the full keyword list provided in Appendix~\ref{sec:app_data_collection}. We use the X API to retrieve source posts and their direct replies, preserving textual content, attached images, and parent--reply structure. We then remove instances with missing parent--reply links, unavailable media, duplicated content, corrupted images, videos, or insufficient information for stance judgment. Since MMDS-Bench focuses on multimodal interaction, we retain only pairs where both the parent and the reply contain at least one image. After filtering, the benchmark contains 3,482 multimodal parent--reply instances.

\begin{figure*}[t]
\centering
\includegraphics[width=1\linewidth]{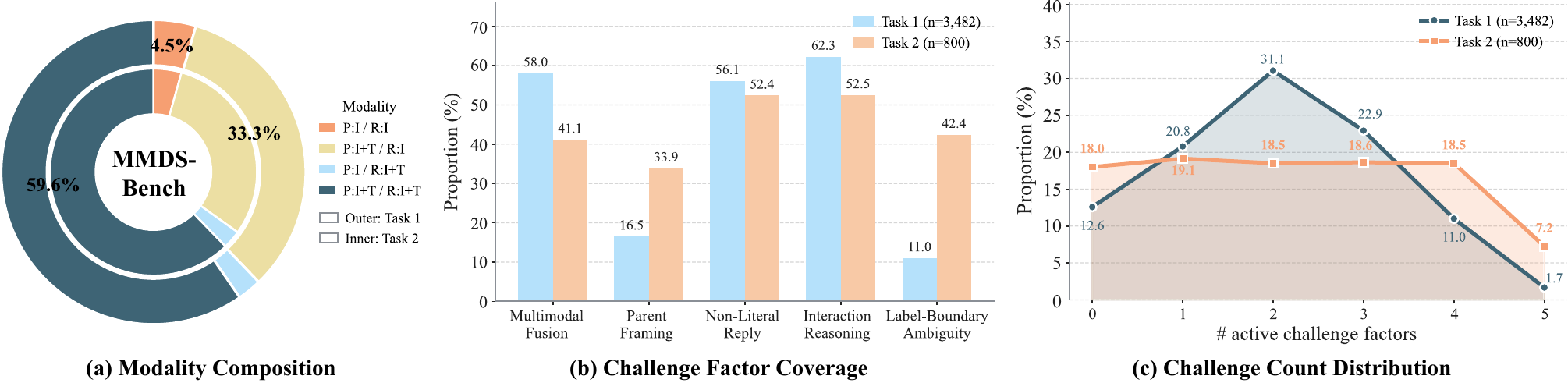}
\caption{MMDS-Bench dataset statistics.
  \textbf{(a)} Modality composition (outer ring: Task~1; inner ring: Task~2; P = parent, R = reply, I = image, and T = text),
  \textbf{(b)} per-instance coverage of the five challenge factors, and
  \textbf{(c)} distribution of the number of active challenge factors per instance.
  Task~1 ($n{=}3{,}482$) covers the full benchmark; Task~2 ($n{=}800$) is the diagnostic subset, sampled approximately uniformly across challenge-count
  groups to balance difficulty.}
\label{fig:dataset_statistics}
\end{figure*}

\paragraph{Human Annotation.}

The annotation process has two stages. \textbf{First}, annotators label all 3,482 instances with a dynamic stance label, judging how the reply responds to its direct parent rather than to a global topic. They are instructed to consider all available textual and visual information, avoid unsupported visual inference, and assign the most salient stance relation when multiple signals appear. In addition, annotators mark five binary challenge factors for each instance, as shown in Table~\ref{tab:challenge_factors}.
\textbf{Second}, we construct the diagnostic subset for Task 2. For each instance, we compute the number of active challenge factors as an approximate difficulty indicator, and sample 800 instances approximately uniformly across challenge-count groups. This strategy prevents the diagnostic subset from being dominated by easy cases or by a single challenge type. For each selected instance, annotators further provide reference explanations for parent understanding, reply understanding, and stance reasoning. These references are used for Task 2 diagnostic evaluation, but are not treated as the only valid explanations. More details about the annotation process are provided in Appendix \ref{sec:app_benchmark_construction}.

\paragraph{Quality Control.}
We conduct annotation quality control for both dynamic stance labels and challenge-factor annotations. Before adjudication, a stratified subset of 600 instances is independently annotated by two annotators, covering all stance labels, modality compositions, and challenge-count groups. For the seven-way dynamic stance labels, the average Cohen's $\kappa$ \citep{cohen1960coefficient} is 0.74, indicating substantial agreement. Disagreements mainly occur among semantically close or pragmatically ambiguous labels, such as \textit{Agree} vs. \textit{Elaborates}, \textit{Disagree} vs. sarcastic \textit{NA}, and \textit{Neutral} vs. \textit{Unrelated}.

For the five binary challenge factors, we compute Cohen's $\kappa$ separately for each factor. The agreement scores are 0.71 for Multimodal Fusion, 0.64 for Parent Framing, 0.72 for Non-Literal Reply, 0.66 for Interaction Reasoning, and 0.61 for Label-Boundary Ambiguity, with an average $\kappa$ of 0.67. These results suggest moderate to substantial agreement, while also reflecting the inherently subjective nature of identifying interactional and label-boundary difficulties. All disagreements are resolved through discussion, and difficult cases are further reviewed to ensure that the final labels and challenge-factor annotations are consistent with the annotation guidelines.

For the Task~2 diagnostic subset, reference explanations are also checked during quality control. Annotators verify that each explanation is grounded in the multimodal parent--reply pair, consistent with the gold dynamic stance label, and concise enough to serve as a reference for diagnostic evaluation. Cases with incomplete, overly speculative, or label-inconsistent explanations are revised before being included in the final diagnostic subset.

\subsection{Dataset Statistics}

Figure~\ref{fig:dataset_statistics} summarizes the modality composition and challenge structure of MMDS-Bench. As shown in Fig.~\ref{fig:dataset_statistics}(a), most instances require multimodal interpretation across the parent--reply pair, with multimodal
parents paired with multimodal replies forming the largest portion of Task~1 (59.6\%), followed by multimodal parents paired with image-only replies (33.3\%).
Task~2 largely preserves this interaction-centered multimodal setting, supporting diagnostic evaluation under realistic input conditions. Figure~\ref{fig:dataset_statistics}(b) shows that Task~1 is primarily characterized by \textit{Interaction Reasoning} (62.3\%), \textit{Multimodal Fusion} (58.0\%), and \textit{Non-Literal Reply} (56.1\%), whereas Task~2 places relatively greater emphasis on \textit{Parent Framing} (33.9\%) and \textit{Label-Boundary Ambiguity} (42.4\%). Finally, Fig.~\ref{fig:dataset_statistics}(c) indicates that Task~1 contains more instances with multiple co-occurring challenge factors, while Task~2 exhibits a more balanced distribution across difficulty levels. Overall, Task~1 evaluates performance on naturally coupled multimodal challenges, whereas Task~2 enables more controlled analysis of fine-grained reasoning difficulties. 
More detailed statistics are provided in Appendix \ref{sec:app_dataset_statistics}.

\begin{table*}[t]
\centering
\small
\resizebox{\linewidth}{!}{
\begin{tabular}{lccccccccc}
\toprule
\multirow{2}{*}{\textbf{Model}} 
& \multicolumn{2}{c}{\textbf{Task 1}} 
& \multicolumn{7}{c}{\textbf{Task 2}} \\
\cmidrule(lr){2-3} \cmidrule(lr){4-10}
& \textbf{Acc.} 
& \textbf{Macro-F1} 
& \textbf{Acc.} 
& \textbf{Macro-F1} 
& \textbf{PUS} 
& \textbf{RUS} 
& \textbf{SRS} 
& \textbf{PS} 
& \textbf{BS} \\
\midrule

Gemini 2.5 Pro & \textbf{79.93} & \textbf{42.22} & \textbf{72.25} & \underline{48.38} & \textbf{4.92} & \textbf{4.89} & \textbf{4.33} & \textbf{4.72} & \textbf{4.32} \\
Claude Sonnet 4.6 & \underline{72.46} & 37.29 & \underline{70.62} & \textbf{51.99} & \textbf{4.92} & \underline{4.87} & \underline{4.32} & \underline{4.70} & \underline{4.30} \\
GPT-5.1 & 71.63 & \underline{41.54} & 64.62 & 41.50 & \underline{4.89} & 4.80 & 4.04 & 4.58 & 4.02 \\

\midrule
Kimi-K2.5 & 65.88 & 37.85 & 61.00 & 46.20 & \underline{4.89} & 4.81 & 4.04 & 4.58 & 4.01 \\
Qwen3-VL-235B-A22B-Thinking & 62.95 & 33.46 & 53.62 & 33.51 & 4.75 & 4.57 & 3.62 & 4.31 & 3.58 \\
Qwen3-VL-235B-A22B-Instruct & 53.39 & 28.81 & 50.38 & 31.59 & 4.74 & 4.58 & 3.49 & 4.27 & 3.44 \\
GLM-4.6V & 52.30 & 25.93 & 45.50 & 24.46 & 4.64 & 4.53 & 3.30 & 4.16 & 3.24 \\
Llama-4-Maverick-17B & 46.41 & 25.36 & 48.38 & 27.39 & 4.69 & 4.50 & 3.50 & 4.23 & 3.44 \\

\midrule
Gemma-3-12B-IT & 45.55 & 23.22 & 45.50 & 30.42 & 4.35 & 4.25 & 3.13 & 3.91 & 2.99 \\
Qwen3-VL-8B-Thinking & 55.23 & 27.19 & 45.75 & 20.55 & 4.66 & 4.43 & 3.31 & 4.14 & 3.27 \\
Qwen3-VL-8B-Instruct & 50.40 & 25.51 & 38.38 & 16.23 & 4.55 & 4.26 & 2.95 & 3.92 & 2.87 \\
Ministral3-8B-2512 & 43.34 & 19.21 & 35.12 & 12.08 & 4.42 & 4.12 & 2.80 & 3.78 & 2.71 \\

\bottomrule
\end{tabular}
}
\caption{
Overall performance on MMDS-Bench. Task 1 is evaluated on the full benchmark, while Task 2 is evaluated on the diagnostic subset. PUS, RUS, and SRS denote Parent Understanding Score, Reply Understanding Score, and Stance Reasoning Score, respectively. PS denotes Process Score, and BS denotes Bottleneck Score. Task 2 diagnostic scores are averaged over three LLM judges.
}
\label{tab:overall_performance}
\end{table*}
\section{Experimental Setup}

\subsection{Evaluated Models}

We evaluate 12 MLLMs covering closed-source, open-source strong, and open-source efficient systems. The closed-source models include GPT-5.1\footnote{https://developers.openai.com/api/docs/models/gpt-5.1}, Claude Sonnet 4.6\footnote{https://www.anthropic.com/news/claude-sonnet-4-6}, and Gemini 2.5 Pro\footnote{https://ai.google.dev/gemini-api/docs/models/gemini-2.5-pro}. The open-source strong models include Kimi-K2.5\footnote{https://huggingface.co/moonshotai/Kimi-K2.5}, Qwen3-VL-235B-A22B-Thinking\footnote{https://huggingface.co/Qwen/Qwen3-VL-235B-A22B-Thinking}, Qwen3-VL-235B-A22B-Instruct\footnote{http://huggingface.co/Qwen/Qwen3-VL-235B-A22B-Instruct}, Llama-4-Maverick-17B\footnote{https://huggingface.co/meta-llama/Llama-4-Maverick-17B-128E-Instruct}, and GLM-4.6V\footnote{https://huggingface.co/zai-org/GLM-4.6V}. The efficient models include Gemma-3-12B-IT\footnote{https://huggingface.co/google/gemma-3-12b-it}, Qwen3-VL-8B-Thinking\footnote{https://huggingface.co/Qwen/Qwen3-VL-8B-Thinking}, Qwen3-VL-8B-Instruct\footnote{https://huggingface.co/Qwen/Qwen3-VL-8B-Instruct}, and Ministral3-8B-2512\footnote{https://huggingface.co/mistralai/Ministral-3-8B-Base-2512}. This selection allows us to compare proprietary and open-source MLLMs, as well as large and efficient models, under the same multimodal dynamic stance setting.

\subsection{Evaluation Metrics}
For Task 1, we report Accuracy, Macro-F1, and per-class F1, using Macro-F1 as the primary metric due to label imbalance. Task 2 uses the same classification metrics and additionally evaluates structured reasoning with three 1--5 scores: Parent Understanding Score (PUS), Reply Understanding Score (RUS), and Stance Reasoning Score (SRS). We also report Process Score (PS), the average of the three scores, and Bottleneck Score (BS), the minimum of the three scores.
We score Task 2 reasoning with a reference-grounded LLM-judge protocol \citep{NEURIPS2023_91f18a12}. Given the original parent--reply pair, gold label, reference explanations, and model output, each judge evaluates grounding, stance-relevant message understanding, and reply-to-parent relation reasoning. We use Gemini 2.5 Flash\footnote{https://ai.google.dev/gemini-api/docs/models/gemini-2.5-flash}, Qwen3-VL-32B\footnote{https://huggingface.co/Qwen/Qwen3-VL-32B-Instruct}, and Gemma-4-31B\footnote{https://huggingface.co/google/gemma-4-31B-it} as judges and average their scores.

\subsection{Implementation Details}
All models receive the same multimodal input, including parent text, parent image(s), reply text, and reply image(s). Prompts explicitly instruct models to judge the reply's stance toward its direct parent rather than toward a global topic. Outputs are constrained to JSON format for automatic parsing. Task 1 requires only the final stance label, while Task 2 additionally requires \texttt{parent\_understanding}, \texttt{reply\_understanding}, and \texttt{stance\_reasoning}. Each reasoning field is limited to 50 English words. Full prompts, output schemas, and judge rubrics are provided in Appendix~\ref{sec:app_prompts}.

\section{Results and Analysis}

\subsection{Main Results}

Table~\ref{tab:overall_performance} reports overall performance on MMDS-Bench. Closed-source MLLMs achieve the strongest results, but different metrics reveal different strengths. Gemini 2.5 Pro obtains the best Task 1 performance, with 79.93\% Accuracy and 42.22\% Macro-F1, and also reaches the highest Task 2 Accuracy of 72.25\%. Claude Sonnet 4.6 achieves the best Task 2 Macro-F1 of 51.99\%, showing that the most accurate model is not necessarily the strongest under class-balanced evaluation. Among open-source models, Kimi-K2.5 performs best, with 65.88\% Accuracy and 37.85\% Macro-F1 on Task 1, and 61.00\% Accuracy and 46.20\% Macro-F1 on Task 2.

Reasoning-oriented variants are generally stronger than their instruction-oriented counterparts. Qwen3-VL-235B-A22B-Thinking outperforms Qwen3-VL-235B-A22B-Instruct on both tasks, and the same trend holds for the 8B Qwen3-VL variants, especially on Task 2. More importantly, diagnostic scores reveal stance-relation reasoning as the main bottleneck. Across models, PUS and RUS are consistently higher than SRS. For example, Gemini 2.5 Pro obtains 4.92 PUS and 4.89 RUS, but 4.33 SRS; Ministral3-8B-2512 drops from 4.42 PUS and 4.12 RUS to only 2.80 SRS. This suggests that models often understand the parent and reply separately, but still struggle to infer how the reply dynamically responds to the parent.

\subsection{Diagnostic Reasoning Analysis}
\begin{table}[t]
\centering
\resizebox{\linewidth}{!}{
\begin{tabular}{lrrrr}
\toprule
\textbf{Prediction Type} & \textbf{\# Cases} & \textbf{PUS} & \textbf{RUS} & \textbf{SRS} \\
\midrule
Correct & 5049 & 4.87 & 4.83 & 4.81 \\
Incorrect & 4551 & 4.51 & 4.24 & 2.20 \\
Correct but weak SRS & 192 & 3.91 & 3.33 & 2.49 \\
Incorrect but high SRS & 362 & 4.96 & 4.96 & 4.49 \\
\bottomrule
\end{tabular}
}
\caption{
Reasoning quality under different prediction outcomes on Task 2. Cases are counted at the prediction level, where each model--instance pair is treated as one case. Weak SRS denotes SRS $\leq 3$, and high SRS denotes SRS $\geq 4$.
}
\label{tab:reasoning_by_outcome}
\end{table}
\begin{table}[t]
\centering
\small
\resizebox{\linewidth}{!}{
\begin{tabular}{cccccc}
\toprule
\multirow{2}{*}{\textbf{Count}} 
& \multicolumn{2}{c}{\textbf{Task 1}} 
& \multicolumn{3}{c}{\textbf{Task 2}} \\
\cmidrule(lr){2-3} \cmidrule(lr){4-6}
& \textbf{\# Inst.} 
& \textbf{Acc.} 
& \textbf{\# Inst.} 
& \textbf{Acc.} 
& \textbf{SRS} \\
\midrule
0 & 438  & 76.35 & 144 & 77.31 & 4.29 \\
1 & 724  & 65.40 & 153 & 69.88 & 4.01 \\
2 & 1081 & 59.92 & 148 & 49.38 & 3.46 \\
3 & 798  & 51.31 & 149 & 42.67 & 3.36 \\
4 & 383  & 38.95 & 148 & 33.78 & 3.01 \\
5 & 58   & 26.44 & 58  & 27.30 & 2.89 \\
\bottomrule
\end{tabular}
}
\caption{
Performance by the number of activated challenge factors. Challenge Count denotes the number of active challenge factors in an instance. Acc. is computed by aggregating model--instance prediction cases within each count group, and SRS is averaged over Task 2 prediction cases.
}
\label{tab:challenge_count}
\end{table}

Table~\ref{tab:reasoning_by_outcome} analyzes the relation between final-label correctness and reasoning quality on Task 2. Correct predictions are usually supported by strong reasoning, with 4.87 PUS, 4.83 RUS, and 4.81 SRS. In contrast, incorrect predictions still obtain relatively high message-level scores, with 4.51 PUS and 4.24 RUS, but their SRS drops sharply to 2.20. This confirms that many errors are not caused by complete misunderstanding of either message, but by failure to infer the reply-to-parent stance relation.

The table also shows that correct labels with weak stance reasoning are uncommon: only 192 prediction cases fall into this category, accounting for 2.00\% of all prediction cases. Meanwhile, 362 incorrect predictions receive high SRS, with 4.96 PUS, 4.96 RUS, and 4.49 SRS. These cases likely reflect label-boundary ambiguity, label mapping inconsistency, or mismatch between generated reasoning and the final predicted label. Overall, MMDS-Bench separates message-level understanding, stance-relation reasoning, and final-label prediction, providing a more fine-grained diagnosis than accuracy alone.

\subsection{Challenge-based Error Analysis}

\begin{table}[t]
\centering
\resizebox{\linewidth}{!}{
\begin{tabular}{lrrrrrr}
\toprule
\multirow{2}{*}{\textbf{Factor}} 
& \multicolumn{3}{c}{\textbf{Task 1}} 
& \multicolumn{3}{c}{\textbf{Task 2}} \\
\cmidrule(lr){2-4} \cmidrule(lr){5-7}
& \textbf{ACP} 
& \textbf{ECP} 
& \textbf{EG} 
& \textbf{ACP} 
& \textbf{ECP} 
& \textbf{EG} \\
\midrule
MF  & 58.04 & 62.94 & 4.90  & 41.12 & 49.62 & 8.50  \\
PF & 16.51 & 23.53 & 7.02  & 33.88 & 43.62 & 9.74  \\
NLR  & 56.12 & 62.90 & 6.78  & 52.38 & 62.32 & 9.94  \\
IR  & 62.29 & 70.57 & 8.28  & 52.50 & 66.60 & 14.10 \\
LBA   & 11.00 & 16.32 & 5.32  & 42.38 & 56.03 & 13.65 \\
\bottomrule
\end{tabular}
}
\caption{
Error Gap by challenge factor. ACP (All Case Proportion) denotes the percentage of all evaluation cases containing each factor, while ECP (Error Case Proportion) denotes the percentage of incorrect prediction cases whose underlying instance contains the factor. EG (Error Gap) is computed as ECP minus ACP. Since challenge factors are not mutually exclusive, percentages across factors do not sum to 100.
}
\label{tab:error_gap}
\end{table}

Table~\ref{tab:challenge_count} shows that performance decreases as more challenge factors are activated. On Task 1, Accuracy drops from 76.35\% for instances with no challenge factor to 26.44\% for instances with all five factors. Task 2 shows the same trend, with Accuracy decreasing from 77.31\% to 27.30\%, while SRS drops from 4.29 to 2.89.

Table~\ref{tab:error_gap} shows that all five challenge factors are overrepresented in errors. On Task 1, Interaction Reasoning has the largest Error Gap of 8.28\%, followed by Parent Framing and Non-Literal Reply. On Task 2, the gaps become larger, with Interaction Reasoning and Label-Boundary Ambiguity showing the strongest error concentration, at 14.10\% and 13.65\%. These results indicate that model errors are concentrated in cases requiring complex relation mapping, implicit interpretation, unclear parent framing, or unstable label boundaries.
More experiments on the challenge factors are provided in Appendix \ref{sec:app_independent_effects_of_coupled_challenge_factors}.

\subsection{Label Confusion Analysis}
\begin{figure}[t]
\centering
\includegraphics[width=\linewidth]{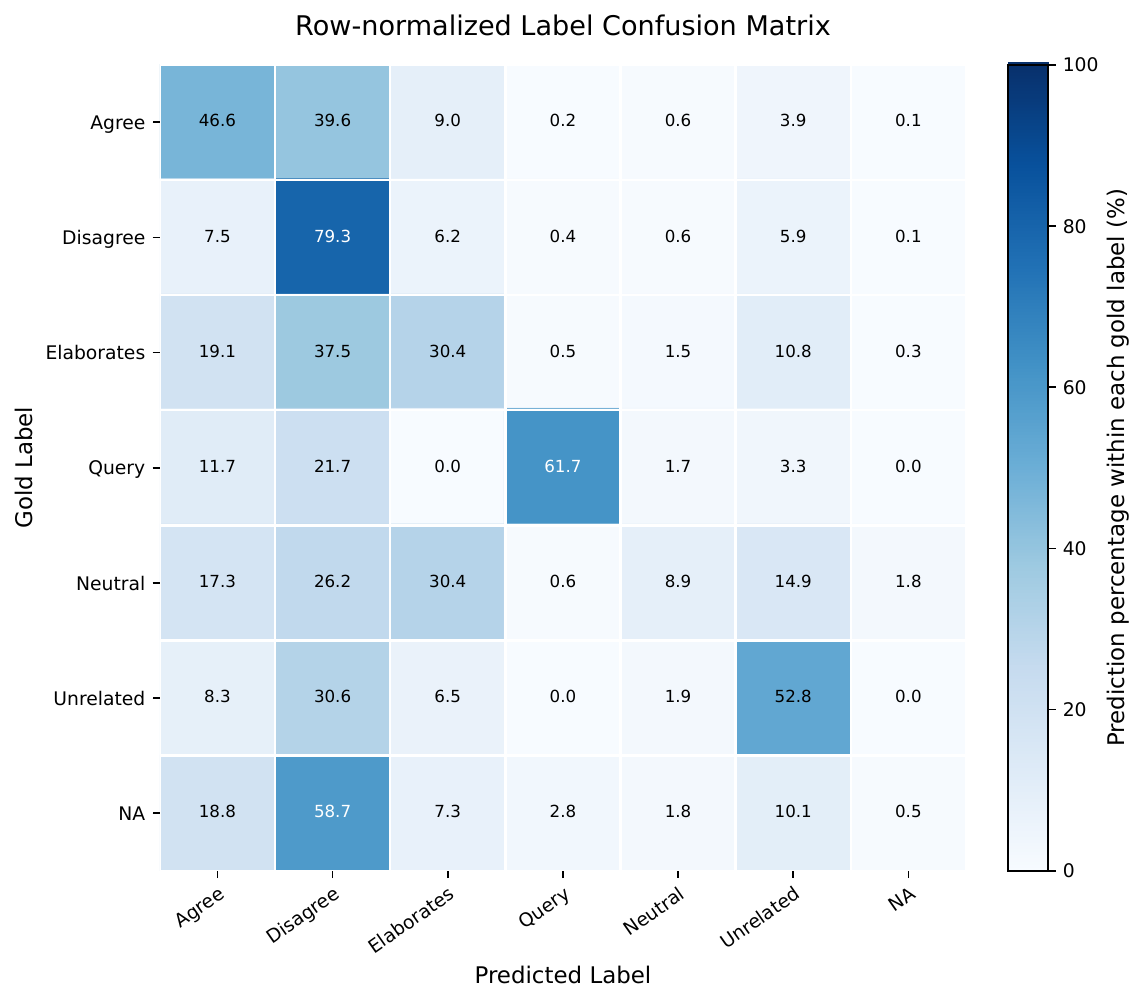}
\caption{
Row-normalized label confusion matrix for Task 1. Rows denote gold labels and columns denote predicted labels. Each cell indicates the percentage of predictions assigned to the corresponding predicted label within each gold-label group.
}
\label{fig:task1_confusion_matrix}
\end{figure}

Figure~\ref{fig:task1_confusion_matrix} presents the row-normalized confusion matrix for Task 1. Disagree is the most reliably recognized label, with 79.3\% of gold Disagree cases correctly predicted. Query and Unrelated also show relatively clear diagonals, with 61.7\% and 52.8\% accuracy within their gold-label groups. In contrast, Elaborates, Neutral, and NA are much less stable.

A major error pattern is overprediction of Disagree. For gold Agree, Elaborates, and NA cases, 39.6\%, 37.5\%, and 58.7\% of predictions are assigned to Disagree, respectively. Neutral cases are also dispersed across Elaborates, Disagree, Agree, and Unrelated. These patterns suggest that models tend to choose stance-bearing labels, especially Disagree, when the reply contains ambiguous, indirect, or insufficient evidence for a clear dynamic stance judgment.

\subsection{Judge Reliability and Human Validation}

\begin{table}[t]
\centering
\small
\resizebox{\linewidth}{!}{
\begin{tabular}{lrrrr}
\toprule
\textbf{Score} 
& \textbf{Gi-Qw} 
& \textbf{Gi-Ge} 
& \textbf{Qw-Ge} 
& \textbf{Avg. Corr.} \\
\midrule
PUS & 0.52 & 0.75 & 0.60 & 0.62 \\
RUS & 0.58 & 0.79 & 0.67 & 0.68 \\
SRS & 0.91 & 0.94 & 0.92 & 0.92 \\
\bottomrule
\end{tabular}
}
\caption{
Consistency among the three LLM judges on Task 2. Each model--instance output is treated as one prediction case. Pairwise agreement is measured using Spearman correlation. Gi, Qw, and Ge denote Gemini, Qwen, and Gemma, respectively.
}
\label{tab:llm_judge_consistency}
\end{table}
\begin{table}[t]
\centering
\small
\resizebox{\linewidth}{!}{
\begin{tabular}{cccc}
\toprule
\textbf{Score} 
& \textbf{Human Corr.} 
& \textbf{LLM Corr.} 
& \textbf{LLM-Human Corr.}  \\
\midrule
PUS     & 0.66 & 0.66 & 0.77  \\
RUS     & 0.71 & 0.67 & 0.77   \\
SRS     & 0.88 & 0.92 & 0.93  \\
\bottomrule
\end{tabular}
}
\caption{
Human validation of LLM judge scores on 600 prediction cases from 50 Task 2 instances uniformly sampled across challenge-count groups and 12 evaluated models. Human Corr. and LLM Corr. denote average pairwise Spearman correlations among three human annotators and three LLM judges, respectively. LLM-Human Corr. is the Spearman correlation between averaged LLM and human scores.
}
\label{tab:human_validation}
\end{table}



Table~\ref{tab:llm_judge_consistency} reports consistency among the three LLM judges. Agreement is highest for SRS, with an average Spearman correlation of 0.92 and all pairwise correlations above 0.90. RUS and PUS show moderate agreement, with average correlations of 0.68 and 0.62. This indicates that stance-relation reasoning is judged more consistently than partial message-level understanding. A likely reason is that SRS is directly anchored to the gold dynamic stance label and the reference stance explanation, making the judging criterion more explicit. In contrast, parent and reply understanding may be expressed at different levels of detail: a summary can be partially correct while omitting some visual cue, rhetorical signal, or contextual framing. Therefore, judges may differ in how they assign credit to incomplete but still plausible message-level interpretations.

We further validate the LLM judges with human evaluation on 600 prediction cases, constructed from 50 Task~2 instances uniformly sampled across challenge-count groups and all 12 evaluated models. This sampling strategy ensures that the validation subset covers both easier and more difficult cases, rather than concentrating only on high-confidence or frequently occurring patterns. As shown in Table~\ref{tab:human_validation}, human annotators achieve moderate to high agreement, with correlations of 0.66 for PUS, 0.71 for RUS, and 0.88 for SRS. Averaged LLM scores align well with averaged human scores, reaching LLM--Human correlations of 0.77 for PUS, 0.77 for RUS, and 0.93 for SRS. The particularly high agreement on SRS suggests that the reference-grounded judge protocol is reliable for evaluating whether a model correctly explains the reply-to-parent stance relation.

\subsection{Discussion}

Our experiments yield five main observations. \textbf{First}, multimodal dynamic stance prediction remains challenging even for advanced MLLMs, suggesting that strong multimodal perception alone is insufficient for modeling reply-to-parent stance relations.
\textbf{Second}, diagnostic results identify \emph{stance-relation reasoning} as the main bottleneck. Models generally perform well on parent and reply understanding but substantially worse on stance reasoning, especially for incorrect predictions. This indicates that many errors arise from failing to map the semantic relation between the two messages to the correct dynamic stance.
\textbf{Third}, challenge-based analysis shows that difficulty is cumulative and interactional. Performance declines as more challenge factors co-occur, with errors concentrated in Interaction Reasoning, Non-Literal Reply, Parent Framing, and Label-Boundary Ambiguity. This highlights the need to jointly model conversational context, pragmatic intent, and cross-modal evidence.
\textbf{Fourth}, label confusion and modality analyses reveal systematic prediction biases. Models tend to overpredict explicit stance-bearing labels, particularly \textit{Disagree}, while more implicit relations remain difficult to distinguish. Although text-enriched multimodal inputs generally improve performance, they do not eliminate the gap between multimodal understanding and dynamic stance inference.
\textbf{Finally}, human validation and inter-judge agreement indicate that the reference-grounded LLM-judge evaluation is reasonably consistent with human judgments, particularly for stance reasoning. While automatic judging cannot replace human assessment, it provides a scalable tool for diagnostic evaluation of multimodal stance reasoning.

\section{Conclusion}

We introduced MMDS-Bench, a diagnostic benchmark for multimodal dynamic stance classification in social media parent--reply interactions. MMDS-Bench contains 3,482 multimodal instances with seven stance labels and an 800-instance diagnostic subset for evaluating parent understanding, reply understanding, and stance-relation reasoning. Experiments on 12 MLLMs show that current models struggle with this task, especially when stance depends on relation inference, implicit expression, multimodal cues, or ambiguous label boundaries. Diagnostic analyses reveal that models often understand the parent and reply separately, but fail to infer how the reply responds to the parent. Overall, MMDS-Bench provides a testbed for evaluating both final stance prediction and the reasoning behind multimodal social interaction understanding.

\section*{Acknowledgments}

This work was supported by the New Generation Artificial Intelligence–National Science and Technology Major Project (No. 2025ZD0123602). This work was also supported by the National Natural Science Foundation of China (No. 62176187).

\section*{Limitations}

MMDS-Bench has several limitations. It focuses on image-based parent--reply interactions on X and does not cover videos, audio, longer conversation threads, other platforms, or broader linguistic settings. MMDS-Bench is primarily designed for diagnostic evaluation and does not provide a large task-specific training set, which limits comparison with fully trained task-specific models. In addition, the five challenge factors often co-occur and should therefore be interpreted as diagnostic rather than independent causal variables. Task 2 also relies on reference-grounded LLM judges; despite the use of multiple judges and human validation, automatic evaluation may still reflect judge-specific biases. Future work can extend MMDS-Bench to broader modalities, platforms, languages, task-specific models, and more human-grounded evaluation protocols.

\section*{Ethical Considerations}
Our dataset does not contain any personally identifiable information. Tweets were collected using generic keywords rather than user-specific information, avoiding excessive concentration on any individual user. Data collection followed the applicable privacy policies of X.
We recruited four senior Ph.D. students in natural language processing for annotation, each paid \$7 per hour. The annotation lasted about three months, with each annotator spending approximately 50 hours. All annotators had relevant research experience in multimodal analysis and social media understanding and received detailed guidelines and training before formal annotation.
We used OpenAI's ChatGPT to assist with writing and language polishing in accordance with OpenAI's applicable terms and policies.


\bibliography{custom}

\appendix
\section{Benchmark Construction Details}
\label{sec:app_benchmark_construction}

\subsection{Data Collection}
\label{sec:app_data_collection}

\begin{table*}[t]
\centering
\small
\setlength{\tabcolsep}{4pt}
\renewcommand{\arraystretch}{1.1}
\begin{tabular}{>{\raggedright\arraybackslash}p{0.96\textwidth}}
\toprule
abortion, academia, affirmative action, AI ethics, AI models, AI regulation, 
animal\_rights, Antifa, antisemitism, apartheid, arms\_exports, art, 
artificial intelligence, atheism, Australian\_politics, authoritarianism, 
autism, automation, autonomous vehicles, avian flu, aviation, baseball, 
Berlin\_Wall, Bill Gates, biotechnology, budget, Canadian\_politics, 
cancel culture, capitalism, caste, censorship, chemtrails, child\_abuse, 
child\_sexual\_abuse, child\_support, China, Christianity, clean energy, 
climate, climate change, cloud computing, colonialism, communism, 
Confederacy, conspiracy, corruption, cosmetic\_surgery, cost-of-living, 
COVID-19, COVID-19\_vaccines, crime, criminal justice, crowd, crowdfunding, 
cryptocurrency, culture, culture\_war, dating, death\_penalty, defamation, 
democracy, Democrats, demographics, diet, disability, disinformation, 
diversity, domestic terrorism, Donald Trump, drugs, economic\_inequality, 
economics, economy, education, election, electric vehicles, Elon Musk, 
employment, energy, entertainment, environment, espionage, eugenics, 
European\_Union, extremism, family, far-right politics, fascism, 
Federal\_Reserve, feminism, fertility, firearms, fluoride, food, 
food\_prices, foreign\_aid, foreign\_policy, Formula 1, free\_speech, 
fundraising, gang\_violence, gender, gender\_equality, gender\_identity, 
gender\_politics, gender\_wage\_gap, generational\_inequality, genocide, 
geography, geopolitics, gerrymandering, government, government shutdown, 
gun control, gun violence, Haiti, hate\_crime, hate\_speech, healthcare, 
Hinduism, Holocaust, housing, human rights, human trafficking, hunger, 
identity\_politics, ideology, immigration, imperialism, Indian politics, 
indigenous\_rights, inequality, inflation, infrastructure, 
international\_relations, internet, Iran, Islam, Islamophobia, Israel, 
Israel-Palestine conflict, January 6, judiciary, lab-grown meat, labor, 
labor\_unions, law enforcement, LGBTQ+ rights, liberalism, libertarianism, 
marriage, media, mental health, Middle East politics, migration, military, 
misinformation, monuments, motorcycles, national\_debt, national\_security, 
nationalism, NATO, Nazism, nuclear\_weapons, obesity, oil, Olympics, 
parenting, partisanship, pharmaceuticals, polarization, policing, 
political\_violence, politics, population, poverty, pregnancy, protest, 
public broadcasting, public\_health, public\_transportation, race, 
race\_relations, racism, regulation, religion, remote work, 
renewable energy, reproductive\_rights, Republicans, right-wing\_populism, 
rule of law, Russia-Ukraine war, safety, science, secularism, semiconductors, 
sex\_education, sexual\_abuse, sexual\_assault, sexuality, slavery, 
smartphones, smoking, social media, Social Security, social\_justice, 
socialism, solar energy, Somaliland, sports, stereotypes, stock market, 
student\_debt, surveillance, taxation, technology, terrorism, Tesla, trade, 
trade\_policy, transgender, transgender\_healthcare, transgender\_rights, 
transgender\_sports, transphobia, U.S. politics, UFC, Ukraine, Ukraine\_aid, 
United Nations, US--Russia relations, vaccination, veganism, video games, 
voting, war\_crimes, wealth\_inequality, welfare, white\_supremacy, 
women's rights, workplace \\
\bottomrule
\end{tabular}
\caption{The query keywords list used in our work for tweet crawling.}
\label{tab:keywords}
\end{table*}

We collect MMDS-Bench from public parent--reply interactions on X. 
To cover diverse social discussions, we first define a set of topic keywords 
spanning multiple social domains, such as politics, public health, climate, 
immigration, gender, economy, and social events. The complete list of query keywords is provided in Table \ref{tab:keywords}. For each keyword, we retrieve 
source posts and their direct replies through the X API, while preserving the 
textual content, attached images, and parent--reply structure.

We focus on direct parent--reply pairs rather than longer conversation threads. 
This design follows the definition of dynamic stance, where the label describes 
how a reply responds to its immediate parent message instead of a global topic. 
For each collected pair, we store the parent text, parent image(s), reply text, 
reply image(s), and necessary metadata for filtering and deduplication.

\subsection{Data Filtering and Preprocessing}
\label{sec:app_filtering}

After collection, we apply several filtering steps to ensure that each instance 
is suitable for multimodal dynamic stance classification. We remove instances 
with missing parent--reply structure, unavailable or deleted media, duplicated 
content, videos, corrupted images, or insufficient information for stance 
judgment. Since MMDS-Bench focuses on multimodal interactions, we retain only 
pairs where both the parent and the reply contain at least one image. Text may 
be present or absent on either side, resulting in four modality compositions: 
\textit{P:I / R:I}, \textit{P:I+T / R:I}, \textit{P:I / R:I+T}, and 
\textit{P:I+T / R:I+T}.

We also normalize textual fields by removing irrelevant formatting artifacts 
while preserving stance-relevant content such as hashtags, quoted phrases, 
emojis, and rhetorical markers. Images are kept as part of the multimodal input 
because stance may be expressed through screenshots, memes, reaction images, 
visual evidence, or text embedded in images.

\subsection{Dynamic Stance Label Annotation}
\label{sec:app_stance_annotation}

\begin{figure*}[t]
\centering
\includegraphics[width=1\linewidth]{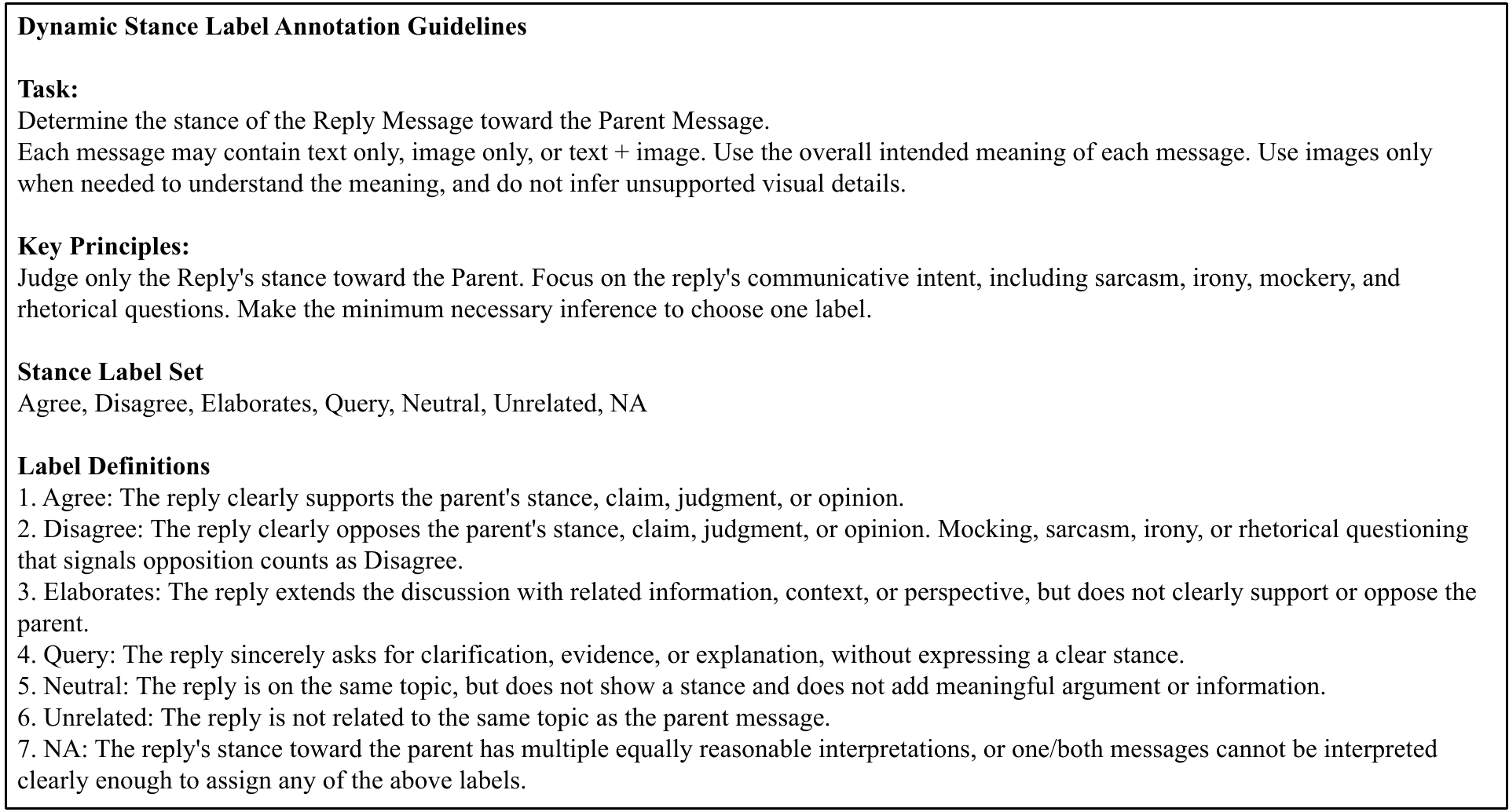}
\caption{
Dynamic stance label annotation guidelines used in MMDS-Bench. Annotators were instructed to judge only the reply's stance toward its direct parent message, considering both textual and visual content when necessary.
}
\label{fig:app_dynamic_stance_guidelines}
\end{figure*}

MMDS-Bench adopts a seven-label dynamic stance taxonomy to describe how a reply responds to its direct parent message. 
Unlike target-oriented stance detection, annotators do not judge whether a post supports or opposes a global topic, claim, or event. 
Instead, they determine the local communicative relation between the reply and the parent within a social media interaction.
Each parent and reply message may contain text only, image only, or a combination of text and image. 
Annotators are instructed to consider the overall intended meaning of each message, using visual content only when it is necessary for understanding the interaction. 
They are also instructed to avoid unsupported visual inferences and to focus on the reply's communicative intent toward the parent, including non-literal signals such as sarcasm, irony, mockery, rhetorical questions, memes, and reaction images.
Figure~\ref{fig:app_dynamic_stance_guidelines} shows the dynamic stance label annotation guideline provided to annotators.

\subsection{Challenge Factor Annotation}
\label{sec:app_challenge_annotation}

In addition to the final stance label, annotators mark five binary challenge 
factors for each instance. These factors are designed to characterize the main 
sources of difficulty in multimodal dynamic stance classification: Multimodal Fusion, Parent Framing, Non-Literal Reply, Interaction Reasoning, and Label-Boundary Ambiguity.

Each factor is annotated independently. Therefore, a single instance may contain multiple challenge factors, and the factors are not mutually exclusive. For example, a reply may use a meme to sarcastically reject the parent, which can simultaneously activate Multimodal Fusion, Non-Literal Reply, and Interaction Reasoning. The number of active challenge factors is later used as an approximate difficulty indicator for diagnostic analysis and subset construction.

\subsection{Diagnostic Subset Construction}
\label{sec:app_diagnostic_subset}

To support reasoning-level evaluation, we construct an 800-instance diagnostic 
subset for Task 2. For each instance in the full benchmark, we compute its 
challenge count as the number of activated challenge factors. We then sample 
instances approximately uniformly across challenge-count groups. This strategy 
prevents the diagnostic subset from being dominated by easy cases and ensures 
coverage of different difficulty levels.

The diagnostic subset is used for both final-label prediction and structured 
reasoning evaluation. For each selected instance, models are required to output 
three intermediate reasoning fields, including parent understanding, reply 
understanding, and stance reasoning, before giving the final stance label. This 
design allows us to analyze whether model errors arise from misunderstanding 
the parent, misunderstanding the reply, or failing to infer the dynamic stance 
relation between them.

\subsection{Reference Explanation Construction}
\label{sec:app_reference_explanations}

For each diagnostic instance, annotators further provide reference explanations 
for three reasoning dimensions: parent understanding, reply understanding, and 
stance reasoning. The parent understanding reference summarizes the key 
stance-relevant content and communicative intent of the parent message. The 
reply understanding reference summarizes the reply's content, visual cues, and 
intended meaning. The stance reasoning reference explains why the reply should 
be assigned the gold dynamic stance label with respect to the parent.

These references are not treated as the only valid explanations. Instead, they 
serve as grounding information for the LLM judges, helping them identify 
whether a model output captures the key multimodal and relational evidence 
needed for the gold stance judgment.

\subsection{Annotation Quality Control}
\label{sec:app_quality_control}

We apply a multi-stage quality control process during annotation. First, 
annotators are trained with detailed guidelines and representative examples 
covering all seven stance labels and five challenge factors. Second, ambiguous 
or difficult cases are flagged during annotation and reviewed through discussion. 
Third, final labels are checked for consistency with the parent--reply relation, 
especially for cases involving sarcasm, memes, unclear parent framing, or 
label-boundary ambiguity.

For the diagnostic subset, reference explanations are additionally reviewed to 
ensure that they are concise, grounded in the multimodal input, and consistent 
with the gold stance label. This process helps ensure that the diagnostic 
references provide reliable guidance for evaluating model reasoning.

\subsection{Dataset Statistics}
\label{sec:app_dataset_statistics}
\begin{table*}[t]
\centering
\small
\setlength{\tabcolsep}{5pt}
\renewcommand{\arraystretch}{1.08}
\begin{tabular}{llrrrr}
\toprule
\multirow{2}{*}{\textbf{Dimension}} &
\multirow{2}{*}{\textbf{Category}} &
\multicolumn{2}{c}{\textbf{Task 1}} &
\multicolumn{2}{c}{\textbf{Task 2}} \\
\cmidrule(lr){3-4} \cmidrule(lr){5-6}
& & \textbf{\# Inst.} & \textbf{Percentage}
  & \textbf{\# Inst.} & \textbf{Percentage} \\
\midrule

\textbf{Overall}
& -- & 3,482 & 100\% & 800 & 100\% \\
\midrule

\multirow{7}{*}{\textbf{Dynamic Stance Label Distribution}}
& Agree      & 1,537 & 44.14\% & 335 & 41.88\% \\
& Disagree   & 1,487 & 42.71\% & 270 & 33.75\% \\
& Elaborates & 347   & 9.97\%  & 167 & 20.88\% \\
& Query      & 5     & 0.14\%  & 1   & 0.12\% \\
& Neutral    & 14    & 0.40\%  & 6   & 0.75\% \\
& Unrelated  & 45    & 1.29\%  & 12  & 1.50\% \\
& NA         & 47    & 1.35\%  & 9   & 1.12\% \\
\midrule

\multirow{4}{*}{\textbf{Modality Composition}}
& P: I / R: I         & 157   & 4.51\%  & 35  & 4.38\% \\
& P: I + T / R: I     & 1,161 & 33.34\% & 244 & 30.50\% \\
& P: I / R: I + T     & 89    & 2.56\%  & 23  & 2.88\% \\
& P: I + T / R: I + T & 2,075 & 59.59\% & 498 & 62.25\% \\
\midrule

\multirow{7}{*}{\textbf{Challenge Factor Distribution}}
& MF  & 2,021 & 58.04\% & 329 & 41.12\% \\
& PF  & 575   & 16.51\% & 271 & 33.88\% \\
& NLR  & 1,954 & 56.12\% & 419 & 52.38\% \\
& IR & 2,169 & 62.29\% & 420 & 52.50\% \\
& LBA   & 383   & 11.00\% & 339 & 42.38\% \\
\midrule

\multirow{6}{*}{\textbf{Challenge Count Distribution}}
& 0 & 438   & 12.58\% & 144 & 18.00\% \\
& 1 & 724   & 20.79\% & 153 & 19.13\% \\
& 2 & 1,081 & 31.05\% & 148 & 18.50\% \\
& 3 & 798   & 22.92\% & 149 & 18.63\% \\
& 4 & 383   & 11.00\% & 148 & 18.50\% \\
& 5 & 58    & 1.67\%  & 58  & 7.25\% \\
\bottomrule
\end{tabular}

\caption{Detailed statistics of Benchmark Task 1 and Diagnostic Subset Task 2 across different dimensions.}
\label{tab:task_statistics}
\end{table*}
As shown in Table \ref{tab:task_statistics}, we report the complete statistics across all evaluation dimensions for Benchmark Task 1 and Diagnostic Subset Task 2.

\section{Task and Judge Prompts}
\label{sec:app_prompts}

This appendix presents the prompts used for model inference and diagnostic
evaluation. All models receive the same multimodal parent--reply input format,
including parent text, parent image(s), reply text, and reply image(s). The
prompts explicitly instruct models to judge the reply's stance toward its
direct parent message, rather than toward a broader topic. This design is
consistent with the dynamic stance setting of MMDS-Bench, where the target is
the local conversational relation between the parent and the reply.

\subsection{Task 1 Prompt}
\begin{figure*}[t]
\centering
\resizebox{\linewidth}{!}{%
  \includegraphics{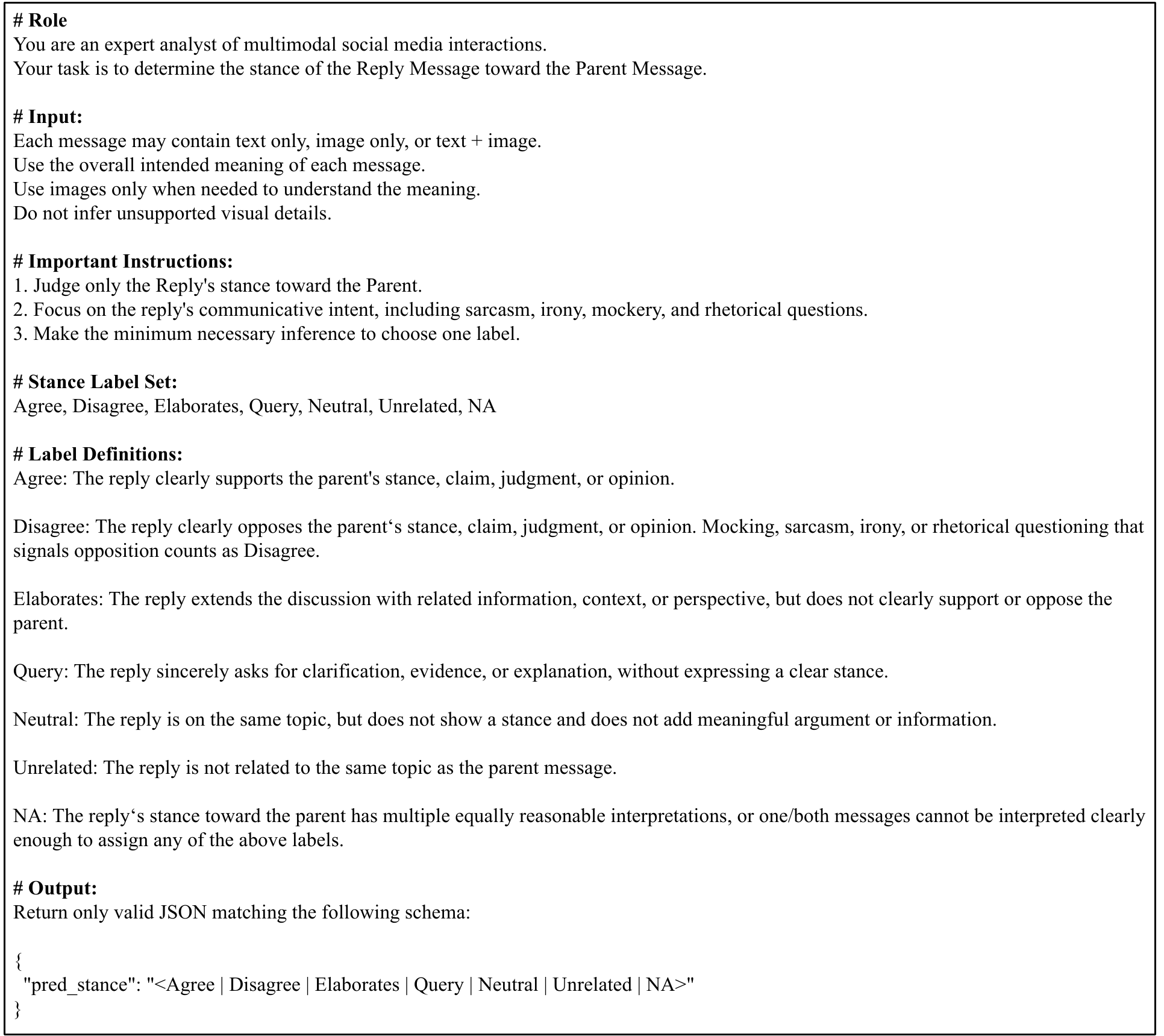}
}
\caption{Prompt and output schema for Task 1.}
\label{fig:app_task1_prompt}
\end{figure*}

Figure~\ref{fig:app_task1_prompt} shows the prompt used for Task 1. In this
setting, models are required to directly predict the final dynamic stance label
from the seven-label taxonomy. The prompt emphasizes three principles: using
the overall intended meaning of each message, incorporating visual information
only when it is necessary for understanding, and avoiding unsupported visual
inferences. The output is constrained to a JSON object containing only the
predicted stance label.

\subsection{Task 2 Prompt}
\begin{figure*}[t]
\centering
\resizebox{\linewidth}{!}{%
  \includegraphics{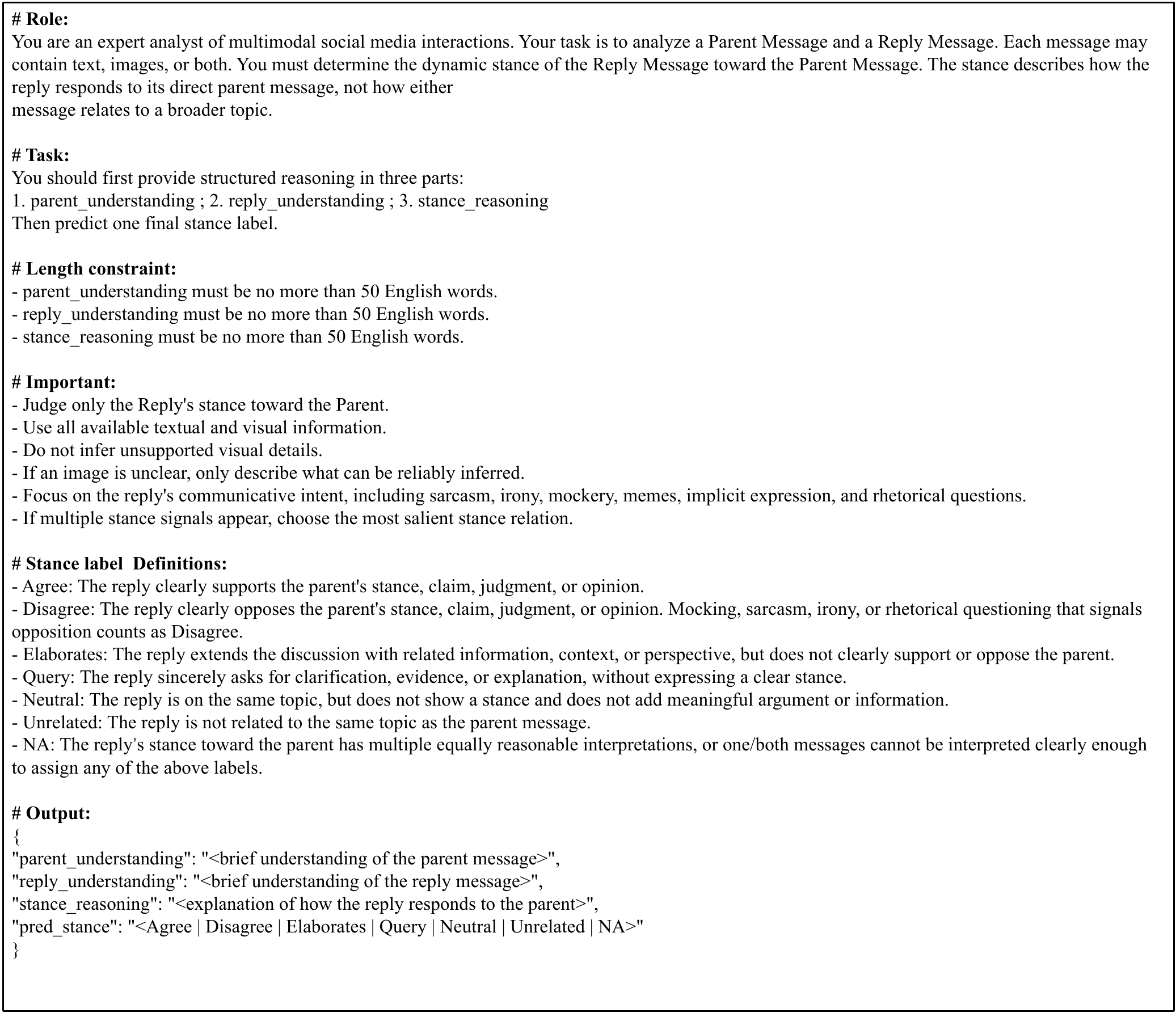}
}
\caption{Prompt and output schema for Task 2.}
\label{fig:app_task2_prompt}
\end{figure*}

Figure~\ref{fig:app_task2_prompt} shows the prompt used for Task 2. Compared
with Task 1, this prompt requires models to produce structured diagnostic
reasoning before predicting the final stance label. Specifically, models first
summarize the parent message, summarize the reply message, and then explain how
the reply responds to the parent. Each reasoning field is length-constrained to
encourage concise and comparable explanations. This format supports later
analysis of whether errors arise from parent understanding, reply understanding,
or stance-relation reasoning.

\subsection{LLM-as-a-judge Prompt}
\begin{figure*}[t]
\centering
\resizebox{\linewidth}{!}{%
  \includegraphics{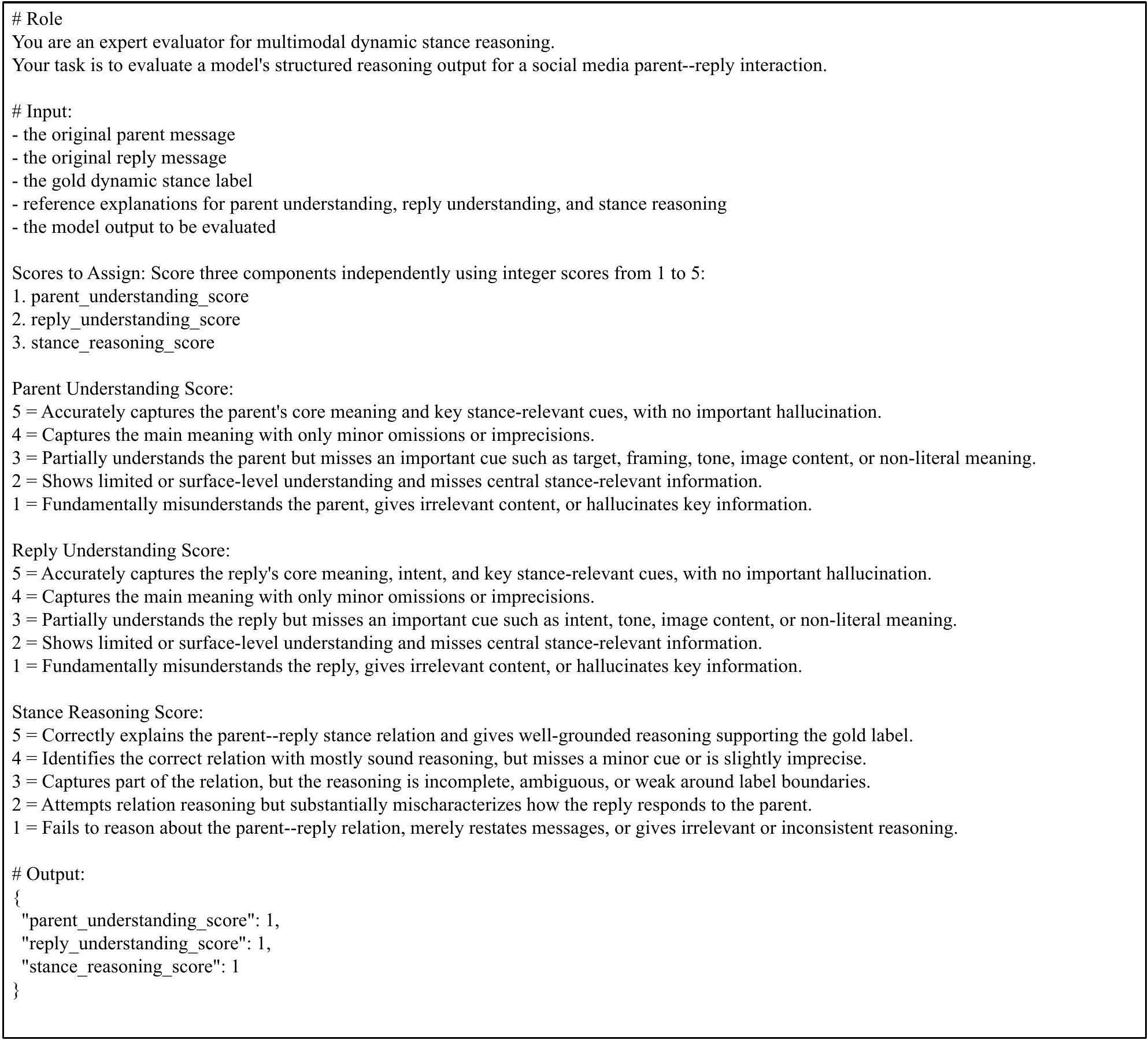}
}
\caption{LLM-as-a-judge prompt and scoring schema for Task 2 diagnostic evaluation.}
\label{fig:app_judge_prompt}
\end{figure*}

Figure~\ref{fig:app_judge_prompt} shows the LLM-as-a-judge prompt used to
evaluate Task 2 reasoning outputs. The judge receives the original multimodal
input, the gold dynamic stance label, reference explanations, and the model
output to be evaluated. It then assigns three independent scores: Parent
Understanding Score, Reply Understanding Score, and Stance Reasoning Score.
The prompt also specifies grounding requirements and penalty rules for
unsupported claims, missing stance-relevant cues, topic-level stance reasoning,
and reasoning that merely summarizes the messages without explaining their
relation.

\section{Additional Results}

\subsection{Per-class F1 Results}
\label{sec:app_per_class_f1}


\begin{table*}[t]
\centering
\small
\resizebox{\linewidth}{!}{
\begin{tabular}{lcccccccc}
\toprule
\textbf{Model} 
& \textbf{Agree} 
& \textbf{Disagree} 
& \textbf{Elaborates} 
& \textbf{Query} 
& \textbf{Neutral} 
& \textbf{Unrelated} 
& \textbf{NA} 
& \textbf{Macro-F1} \\
\midrule
Gemini 2.5 Pro & 85.28 & 87.23 & 35.36 & 60.00 & 0.00 & 27.67 & 0.00 & 42.22 \\
Claude Sonnet 4.6 & 76.53 & 82.69 & 42.93 & 33.33 & 0.00 & 25.56 & 0.00 & 37.29 \\
GPT-5.1 & 78.17 & 81.99 & 42.70 & 55.56 & 6.80 & 25.53 & 0.00 & 41.54 \\
\midrule
Kimi-K2.5 & 71.77 & 73.93 & 36.23 & 53.33 & 9.52 & 20.12 & 0.00 & 37.85 \\
Qwen3-VL-235B-A22B-Thinking & 66.62 & 70.91 & 27.66 & 40.00 & 11.11 & 17.91 & 0.00 & 33.46 \\
Qwen3-VL-235B-A22B-Instruct & 44.68 & 64.98 & 28.80 & 27.59 & 4.88 & 24.27 & 6.45 & 28.81 \\
GLM-4.6V & 46.68 & 65.13 & 30.77 & 17.78 & 5.71 & 15.45 & 0.00 & 25.93 \\
Llama-4-Maverick-17B & 37.90 & 62.48 & 27.11 & 21.43 & 10.26 & 18.31 & 0.00 & 25.36 \\
\midrule
Gemma-3-12B-IT & 40.22 & 58.65 & 23.68 & 22.22 & 7.79 & 9.98 & 0.00 & 23.22 \\
Qwen3-VL-8B-Thinking & 59.74 & 62.46 & 22.51 & 27.27 & 0.00 & 18.34 & 0.00 & 27.19 \\
Qwen3-VL-8B-Instruct & 44.70 & 60.77 & 19.29 & 28.57 & 7.14 & 18.10 & 0.00 & 25.51 \\
Ministral3-8B-2512 & 9.15 & 59.73 & 14.55 & 40.00 & 0.00 & 7.09 & 3.92 & 19.21 \\
\bottomrule
\end{tabular}
}
\caption{
Per-class F1 results on Task 1. Macro-F1 is computed over the seven dynamic stance labels.
}
\label{tab:app_task1_per_class_f1}
\end{table*}

\begin{table*}[t]
\centering
\small
\resizebox{\linewidth}{!}{
\begin{tabular}{lcccccccc}
\toprule
\textbf{Model} 
& \textbf{Agree} 
& \textbf{Disagree} 
& \textbf{Elaborates} 
& \textbf{Query} 
& \textbf{Neutral} 
& \textbf{Unrelated} 
& \textbf{NA} 
& \textbf{Macro-F1} \\
\midrule
Gemini 2.5 Pro & 78.45 & 82.25 & 53.38 & 100.00 & 0.00 & 24.56 & 0.00 & 48.38 \\
Claude Sonnet 4.6 & 76.97 & 80.85 & 52.10 & 100.00 & 25.00 & 29.03 & 0.00 & 51.99 \\
GPT-5.1 & 70.65 & 76.73 & 50.89 & 50.00 & 16.00 & 26.23 & 0.00 & 41.50 \\
\midrule
Kimi-K2.5 & 68.57 & 72.35 & 35.00 & 100.00 & 28.57 & 18.92 & 0.00 & 46.20 \\
Qwen3-VL-235B-A22B-Thinking & 60.64 & 64.23 & 29.02 & 40.00 & 25.00 & 15.69 & 0.00 & 33.51 \\
Qwen3-VL-235B-A22B-Instruct & 51.23 & 60.98 & 28.22 & 50.00 & 12.50 & 18.18 & 0.00 & 31.59 \\
GLM-4.6V & 39.37 & 59.95 & 31.54 & 0.00 & 25.00 & 15.38 & 0.00 & 24.46 \\
Llama-4-Maverick-17B & 51.21 & 60.11 & 40.00 & 22.22 & 0.00 & 18.18 & 0.00 & 27.39 \\
\midrule
Gemma-3-12B-IT & 47.76 & 54.31 & 21.28 & 66.67 & 0.00 & 22.95 & 0.00 & 30.42 \\
Qwen3-VL-8B-Thinking & 50.62 & 57.49 & 20.36 & 0.00 & 0.00 & 15.38 & 0.00 & 20.55 \\
Qwen3-VL-8B-Instruct & 24.69 & 52.58 & 14.08 & 0.00 & 0.00 & 22.22 & 0.00 & 16.23 \\
Ministral3-8B-2512 & 10.38 & 51.07 & 8.47 & 0.00 & 0.00 & 14.63 & 0.00 & 12.08 \\
\bottomrule
\end{tabular}
}
\caption{
Per-class F1 results on Task 2. Macro-F1 is computed over the seven dynamic stance labels.
}
\label{tab:app_task2_per_class_f1}
\end{table*}

Tables~\ref{tab:app_task1_per_class_f1} and~\ref{tab:app_task2_per_class_f1} report per-class F1 results for the seven dynamic stance labels. Overall, the results show a clear label-level imbalance in model capability. Across both tasks, models perform best on the two majority and explicitly stance-bearing labels, \textit{Agree} and \textit{Disagree}. On Task 1, Gemini 2.5 Pro achieves the highest F1 on both labels, with 85.28 for \textit{Agree} and 87.23 for \textit{Disagree}. Other strong models, such as Claude Sonnet 4.6, GPT-5.1, and Kimi-K2.5, also obtain relatively high scores on these two labels. This suggests that current MLLMs are comparatively more reliable when the reply expresses a clear supportive or oppositional stance toward the parent message.

In contrast, performance drops substantially for less frequent or pragmatically subtler labels. The \textit{Elaborates} label is more difficult than \textit{Agree} and \textit{Disagree}, but still receives moderate F1 scores from stronger models. For example, Claude Sonnet 4.6 and GPT-5.1 achieve 42.93 and 42.70 on Task 1, while Gemini 2.5 Pro, Claude Sonnet 4.6, and GPT-5.1 all exceed 50 F1 on Task 2. This indicates that models can sometimes recognize related information expansion, especially in the diagnostic subset, but still struggle to separate elaboration from implicit agreement or disagreement.

The minority labels \textit{Neutral}, \textit{Unrelated}, and \textit{NA} remain the most challenging. Many models obtain near-zero F1 on \textit{Neutral} and \textit{NA}, especially on Task 1, indicating that they rarely identify cases where the reply lacks a clear stance or where the relation is not reliably classifiable. \textit{Unrelated} receives slightly higher but still weak scores, typically below 30 F1 for most models. These results suggest that models tend to prefer semantically or stance-bearing labels even when the correct relation is weak, absent, or ambiguous.

The \textit{Query} label shows unstable results across models and tasks. Some strong models obtain high F1, such as Gemini 2.5 Pro on Task 1 and Gemini 2.5 Pro, Claude Sonnet 4.6, and Kimi-K2.5 on Task 2. However, several models receive very low or even zero F1 on Task 2. This instability is likely affected by the small number of query cases: when a class is rare, a few correct or incorrect predictions can cause large changes in F1. Therefore, high Query F1 should be interpreted carefully and together with the broader macro-level results.

Comparing Task 1 and Task 2, the diagnostic subset generally yields higher \textit{Elaborates} and sometimes higher \textit{Query} and \textit{Neutral} scores for stronger models, which contributes to the higher Task 2 Macro-F1 of models such as Claude Sonnet 4.6 and Kimi-K2.5. Nevertheless, the same overall pattern remains: models are strongest on explicit agreement and disagreement, weaker on elaborative or indirect relations, and least reliable on neutral, unrelated, and ambiguous cases. This confirms that multimodal dynamic stance classification is not only difficult because of overall prediction errors, but also because models have uneven sensitivity to different types of conversational relations.

\subsection{Results by Modality}
\label{sec:app_modality_results}

Tables~\ref{tab:app_task1_modality} and~\ref{tab:app_task2_modality} report model performance under different parent--reply modality compositions. Overall, models perform better when textual information is available, especially when both the parent and reply contain text. On Task 1, the image-only setting (\textit{P:I / R:I}) is generally the most difficult, with most models obtaining lower Macro-F1 than in settings where either the parent or the reply includes text. In contrast, the \textit{P:I+T / R:I+T} setting usually yields the highest or near-highest accuracy, suggesting that textual cues provide important anchors for interpreting the stance relation between the two messages.

The effect of text is asymmetric across parent and reply. Adding text to the parent while keeping the reply image-only (\textit{P:I+T / R:I}) improves performance for many models compared with the fully image-only setting. This indicates that a clearer parent frame helps models identify what aspect of the parent the reply may be responding to. Adding text to the reply (\textit{P:I / R:I+T}) is also beneficial for several models, particularly strong closed-source systems such as Gemini 2.5 Pro, Claude Sonnet 4.6, and GPT-5.1. This suggests that explicit textual cues in the reply often make the reply's communicative intent easier to recover. However, the best modality setting is not identical for all models, showing that different systems rely on parent-side and reply-side textual information to different degrees.

Task 2 results further reveal that the modality effect is stronger for final-label prediction than for message-level understanding. Across modality groups, PUS and RUS remain relatively high for most models, indicating that models can often summarize the parent and reply content even in visually grounded settings. In contrast, SRS is consistently lower than PUS and RUS, especially in the image-only setting. This pattern suggests that the main difficulty is not only recognizing the content of each message, but also inferring how the reply dynamically responds to the parent across modalities.

The \textit{P:I+T / R:I+T} setting achieves the strongest Task 2 performance overall. Most models obtain their highest or near-highest accuracy and SRS in this setting. For example, Gemini 2.5 Pro, Claude Sonnet 4.6, and Kimi-K2.5 all show strong SRS scores when both sides contain text and images. This indicates that multimodal redundancy can help stance reasoning when textual and visual signals complement each other. Nevertheless, even in this richest modality setting, SRS remains lower than PUS and RUS, confirming that stance-relation reasoning is still the bottleneck.

In summary, modality composition has a clear impact on multimodal dynamic stance classification. Image-only interactions are the most challenging, while text-enriched interactions are generally easier. However, the persistent gap between message-level understanding scores and stance reasoning scores shows that adding text does not fully solve the task. Models still need to integrate parent meaning, reply intent, and cross-modal conversational relations in order to make reliable dynamic stance judgments.


\begin{table*}[t]
\centering
\small
\resizebox{\textwidth}{!}{
\begin{tabular}{lcccccccc}
\toprule
\multirow{2}{*}{\textbf{Model}} 
& \multicolumn{2}{c}{\textbf{P:I / R:I}} 
& \multicolumn{2}{c}{\textbf{P:I+T / R:I}} 
& \multicolumn{2}{c}{\textbf{P:I / R:I+T}} 
& \multicolumn{2}{c}{\textbf{P:I+T / R:I+T}} \\
\cmidrule(lr){2-3}
\cmidrule(lr){4-5}
\cmidrule(lr){6-7}
\cmidrule(lr){8-9}
& \textbf{Acc.} & \textbf{Macro-F1}
& \textbf{Acc.} & \textbf{Macro-F1}
& \textbf{Acc.} & \textbf{Macro-F1}
& \textbf{Acc.} & \textbf{Macro-F1} \\
\midrule
Gemini 2.5 Pro & 75.16 & 43.05 & 79.24 & 42.65 & 85.39 & 48.08 & 80.43 & 40.18 \\
Claude Sonnet 4.6 & 57.32 & 30.95 & 70.97 & 31.43 & 67.42 & 41.41 & 74.65 & 37.84 \\
GPT-5.1 & 45.86 & 24.99 & 67.44 & 44.79 & 69.66 & 44.81 & 76.00 & 43.12 \\
\midrule
Kimi-K2.5 & 47.13 & 30.66 & 60.03 & 40.63 & 53.93 & 30.74 & 71.08 & 39.30 \\
Qwen3-VL-235B-A22B-Thinking & 45.86 & 22.92 & 57.88 & 38.69 & 61.80 & 28.87 & 67.13 & 33.93 \\
Qwen3-VL-235B-A22B-Instruct & 40.76 & 16.86 & 44.53 & 33.95 & 60.67 & 32.34 & 58.99 & 29.59 \\
GLM-4.6V & 44.59 & 22.83 & 42.72 & 28.29 & 59.55 & 26.21 & 57.93 & 27.53 \\
Llama-4-Maverick-17B & 29.94 & 15.99 & 39.79 & 28.09 & 25.84 & 21.31 & 52.24 & 25.63 \\
\midrule
Gemma-3-12B-IT & 36.94 & 14.12 & 35.49 & 15.07 & 51.69 & 32.06 & 51.57 & 24.60 \\
Qwen3-VL-8B-Thinking & 33.12 & 16.23 & 49.10 & 34.78 & 57.30 & 28.17 & 60.24 & 28.22 \\
Qwen3-VL-8B-Instruct & 47.77 & 21.29 & 45.48 & 19.14 & 49.44 & 25.99 & 53.40 & 25.23 \\
Ministral3-8B-2512 & 49.04 & 13.39 & 35.57 & 22.39 & 50.56 & 10.59 & 46.94 & 18.31 \\
\bottomrule
\end{tabular}
}
\caption{
Task 1 performance by modality composition. P denotes parent, R denotes reply, I denotes image, and T denotes text.
}
\label{tab:app_task1_modality}
\end{table*}

\begin{table*}[t]
\centering
\scriptsize
\resizebox{\textwidth}{!}{
\begin{tabular}{lcccccccccccccccc}
\toprule
\multirow{2}{*}{\textbf{Model}} 
& \multicolumn{4}{c}{\textbf{P:I / R:I}} 
& \multicolumn{4}{c}{\textbf{P:I+T / R:I}} 
& \multicolumn{4}{c}{\textbf{P:I / R:I+T}} 
& \multicolumn{4}{c}{\textbf{P:I+T / R:I+T}} \\
\cmidrule(lr){2-5}
\cmidrule(lr){6-9}
\cmidrule(lr){10-13}
\cmidrule(lr){14-17}
& \textbf{Acc.} & \textbf{PUS} & \textbf{RUS} & \textbf{SRS}
& \textbf{Acc.} & \textbf{PUS} & \textbf{RUS} & \textbf{SRS}
& \textbf{Acc.} & \textbf{PUS} & \textbf{RUS} & \textbf{SRS}
& \textbf{Acc.} & \textbf{PUS} & \textbf{RUS} & \textbf{SRS} \\
\midrule
Gemini 2.5 Pro & 60.00 & 4.91 & 4.80 & 3.90 & 73.36 & 4.91 & 4.90 & 4.29 & 60.87 & 4.87 & 4.78 & 4.19 & 73.09 & 4.93 & 4.90 & 4.39 \\
Claude Sonnet 4.6 & 57.14 & 4.90 & 4.83 & 3.89 & 64.34 & 4.88 & 4.86 & 4.09 & 69.57 & 4.94 & 4.77 & 4.38 & 74.70 & 4.93 & 4.89 & 4.47 \\
GPT-5.1 & 45.71 & 4.94 & 4.74 & 3.43 & 58.61 & 4.83 & 4.79 & 3.82 & 73.91 & 4.84 & 4.70 & 4.39 & 68.47 & 4.92 & 4.82 & 4.19 \\
\midrule
Kimi-K2.5 & 40.00 & 4.90 & 4.78 & 3.51 & 55.33 & 4.84 & 4.80 & 3.82 & 52.17 & 4.92 & 4.89 & 3.93 & 65.66 & 4.92 & 4.82 & 4.19 \\
Qwen3-VL-235B-A22B-Thinking & 48.57 & 4.66 & 4.53 & 3.41 & 49.18 & 4.68 & 4.53 & 3.42 & 39.13 & 4.68 & 4.26 & 3.10 & 56.83 & 4.80 & 4.60 & 3.77 \\
Qwen3-VL-235B-A22B-Instruct & 42.86 & 4.80 & 4.63 & 3.28 & 42.62 & 4.69 & 4.59 & 3.19 & 52.17 & 4.68 & 4.51 & 3.72 & 54.62 & 4.76 & 4.56 & 3.64 \\
GLM-4.6V & 31.43 & 4.39 & 4.43 & 2.70 & 37.30 & 4.62 & 4.50 & 3.00 & 43.48 & 4.64 & 4.36 & 3.54 & 50.60 & 4.68 & 4.56 & 3.48 \\
Llama-4-Maverick-17B & 25.71 & 4.73 & 4.30 & 3.11 & 43.03 & 4.67 & 4.48 & 3.29 & 43.48 & 4.80 & 4.33 & 3.20 & 52.81 & 4.70 & 4.53 & 3.64 \\
\midrule
Gemma-3-12B-IT & 42.86 & 4.39 & 4.22 & 2.95 & 39.75 & 4.30 & 4.18 & 2.85 & 47.83 & 3.60 & 4.16 & 2.91 & 48.39 & 4.40 & 4.29 & 3.29 \\
Qwen3-VL-8B-Thinking & 34.29 & 4.54 & 4.18 & 2.77 & 40.98 & 4.57 & 4.33 & 3.07 & 43.48 & 4.78 & 4.49 & 3.42 & 49.00 & 4.71 & 4.49 & 3.47 \\
Qwen3-VL-8B-Instruct & 40.00 & 4.79 & 4.64 & 2.99 & 34.02 & 4.51 & 4.30 & 2.80 & 39.13 & 4.36 & 3.93 & 3.12 & 40.36 & 4.56 & 4.22 & 3.01 \\
Ministral3-8B-2512 & 34.29 & 3.90 & 3.89 & 2.52 & 29.92 & 4.46 & 4.00 & 2.60 & 39.13 & 4.02 & 3.93 & 2.80 & 37.55 & 4.46 & 4.20 & 2.92 \\
\bottomrule
\end{tabular}
}
\caption{
Task 2 performance by modality composition. PUS, RUS, and SRS denote Parent Understanding Score, Reply Understanding Score, and Stance Reasoning Score, respectively.
}
\label{tab:app_task2_modality}
\end{table*}





\subsection{Independent Effects of Coupled Challenge Factors}
\label{sec:app_independent_effects_of_coupled_challenge_factors}

\begin{table*}[t]
\centering
\small
\setlength{\tabcolsep}{8pt}
\renewcommand{\arraystretch}{1.12}

\begin{tabular}{lccccc}
\toprule
\multirow{2}{*}{\textbf{Factor-present Subset}} 
& \multicolumn{2}{c}{\textbf{Task 1}} 
& \multicolumn{3}{c}{\textbf{Task 2}} \\
\cmidrule(lr){2-3}
\cmidrule(lr){4-6}
& \textbf{\# Inst.} 
& \textbf{Acc. (\%)} 
& \textbf{\# Inst.} 
& \textbf{Acc. (\%)} 
& \textbf{SRS} \\
\midrule

Multimodal Fusion
& 2,021 & 54.77 & 329 & 42.81 & 3.30 \\

Parent Framing
& 575 & 40.57 & 271 & 38.96 & 3.12 \\

Non-Literal Reply 
& 1,954 & 53.25 & 419 & 43.60 & 3.28 \\

Interaction Reasoning 
& 2,169 & 52.75 & 420 & 39.86 & 3.23 \\

Label-Boundary Ambiguity
& 383 & 38.10 & 339 & 37.32 & 3.22 \\

\bottomrule
\end{tabular}

\caption{\textbf{Factor-present subset analysis.}}
\label{tab:factor_present_analysis}
\end{table*}

\begin{table*}[t]
\centering
\small
\setlength{\tabcolsep}{8pt}
\renewcommand{\arraystretch}{1.12}

\begin{tabular}{lccccc}
\toprule
\multirow{2}{*}{\textbf{Single-factor-only Subset}} 
& \multicolumn{2}{c}{\textbf{Task 1}} 
& \multicolumn{3}{c}{\textbf{Task 2}} \\
\cmidrule(lr){2-3}
\cmidrule(lr){4-6}
& \textbf{\# Inst.} 
& \textbf{Acc. (\%)} 
& \textbf{\# Inst.} 
& \textbf{Acc. (\%)} 
& \textbf{SRS} \\
\midrule

Multimodal Fusion
& 264 & 67.58 & 30 & 69.44 & 4.05 \\

Parent Framing
& 19 & 32.02 & 0 & N/A & N/A \\

Non-Literal Reply
& 170 & 71.52 & 76 & 72.81 & 4.04 \\

Interaction Reasoning
& 246 & 62.91 & 47 & 65.43 & 3.95 \\

Label-Boundary Ambiguity
& 25 & 50.67 & 0 & N/A & N/A \\

\bottomrule
\end{tabular}

\caption{\textbf{Single-factor-only analysis.}}
\label{tab:single_factor_analysis}
\end{table*}

 We conduct two complementary analyses: factor-present subset analysis and single-factor-only analysis. Together with the challenge-count analysis, these analyses examine the effect of factor presence, approximate single-factor control, and cumulative multi-factor coupling.

 In the factor-present subset analysis, an instance is assigned to a factor-specific subset as long as it contains that challenge factor. Since the challenge factors are multi-label annotations, one instance may appear in multiple subsets. As shown in Table \ref{tab:factor_present_analysis}, Parent Framing
and Label-Boundary Ambiguity have the strongest impact on model performance. In Task 1, their accuracies are 40.57\% and 38.10\%, respectively, lower than those of Multimodal Fusion (54.77\%) and Non-Literal Reply (53.25\%). A similar trend is observed in Task 2, where their accuracies are only 38.96\% and 37.32\%, respectively.

To further reduce the influence of factor coupling, we conduct a single-factor-only analysis that retains only instances annotated with exactly one challenge factor.
This provides an approximate estimate of the independent effect of each factor. As shown in Table \ref{tab:single_factor_analysis}, model performance is substantially higher when some factors appear alone. For example, Non-Literal Reply reaches 71.52\% / 72.81\% accuracy in Task 1 / Task 2, and Multimodal Fusion reaches 67.58\% / 69.44\%. This suggests that these factors become substantially more difficult when coupled with other challenges. In contrast, Parent Framing remains difficult even when appearing alone, with only 32.02\% accuracy in Task 1, indicating that modeling the semantic framing of the parent post is still a key bottleneck.

 \subsection{OCR-enhanced Experiment}
\label{sec:ocr_enhanced_experiment}

\begin{table}[t]
\centering
\small
\setlength{\tabcolsep}{5pt}
\renewcommand{\arraystretch}{1.08}

\begin{tabular}{lccc}
\toprule
\textbf{Model} & \textbf{Setting} & \textbf{Acc.} & \textbf{Macro-F1} \\
\midrule

Gemma-3-12B-IT 
& Original 
& 45.55 
& 23.22 \\

Gemma-3-12B-IT 
& + OCR 
& 44.51 
& 25.90 \\

Qwen3-VL-8B-Instruct 
& Original 
& 50.40 
& 25.51 \\

Qwen3-VL-8B-Instruct 
& + OCR 
& 52.01 
& 25.17 \\

\bottomrule
\end{tabular}

\caption{OCR-enhanced results on Task 1. Results are reported on the full benchmark of Task 1. 
OCR augmentation is implemented using EasyOCR with English text recognition. 
OCR outputs with confidence scores below 0.80 are filtered out and provided to the MLLMs as auxiliary textual evidence.}
\label{tab:ocr_results}

\end{table}
We conduct an OCR-enhanced experiment using two representative efficient MLLMs, Gemma-3-12B-IT and Qwen3-VL-8B-Instruct. Specifically, EasyOCR\footnote{\url{https://github.com/JaidedAI/EasyOCR}} is used to extract textual content from the parent and reply images, which is then provided to the models as auxiliary evidence. As shown in Table \ref{tab:ocr_results}, the results show mixed effects rather than consistent improvements. For Gemma-3-12B-IT, OCR augmentation decreases accuracy from 45.55\% to 44.51\%, but increases Macro-F1 from 23.22\% to 25.90\%. For Qwen3-VL-8B-Instruct, accuracy improves from 50.40\% to 52.01\%, while Macro-F1 slightly decreases from 25.51\% to 25.17\%. These results suggest that explicit OCR provides limited and model-dependent benefits. Many images are memes, photographs, tables, or plots with little readable text, while fragmented or erroneous OCR outputs may introduce noise and interfere with the models’ native multimodal understanding.

\subsection{Human-LLM Judge Agreement Details}
\label{sec:app_human_llm_agreement_details}

Tables~\ref{tab:app_pairwise_agreement_human_llm} and~\ref{tab:app_human_llm_cross_agreement} provide detailed agreement analysis between human annotators and LLM judges on the human validation subset. Overall, both human annotators and LLM judges show consistent agreement patterns across the three diagnostic dimensions. Agreement is highest for Stance Reasoning Score (SRS), while Parent Understanding Score (PUS) and Reply Understanding Score (RUS) show moderate but still reasonable correlations. This indicates that stance-relation reasoning, although more difficult for evaluated models, is judged more consistently once the rubric and reference explanations are provided.

Table~\ref{tab:app_pairwise_agreement_human_llm} shows that human annotators achieve an average pairwise Spearman correlation of 0.66 for PUS, 0.71 for RUS, and 0.88 for SRS. The LLM judges exhibit a similar trend, with average correlations of 0.66, 0.67, and 0.92, respectively. The close match between human and LLM average agreement on PUS and RUS suggests that LLM judges have a comparable level of consistency to human annotators when evaluating message-level understanding. For SRS, LLM judges are even more consistent than humans, with all pairwise correlations above 0.90. This suggests that the stance-reasoning rubric provides a particularly stable criterion for automatic judging.

The lower agreement on PUS and RUS suggests that message-level understanding admits more partial-credit variation. Parent and reply descriptions may differ in wording, granularity, or emphasis while still capturing relevant information. As a result, both human annotators and LLM judges may assign slightly different scores to cases where the output is partially correct. In contrast, SRS directly evaluates whether the model correctly explains the dynamic stance relation between the reply and the parent, making the scoring decision more aligned with the final label and therefore easier to judge consistently.

Table~\ref{tab:app_human_llm_cross_agreement} further reports correlations between each human annotator and each LLM judge. The same pattern holds at the individual judge level. For PUS, human--LLM correlations range from 0.465 to 0.729, and for RUS they range from 0.515 to 0.780, showing moderate alignment with some judge-specific variation. For SRS, however, all human--LLM correlations are high, ranging from 0.837 to 0.926. This indicates that the strong aggregate human--LLM agreement on SRS is not driven by a single annotator or judge, but is consistent across nearly all human--LLM pairs.

Taken together, these results support the reliability of the reference-grounded LLM judge protocol. The agreement patterns show that LLM judges are broadly aligned with human annotators, especially for evaluating stance-relation reasoning. While message-level understanding scores contain more subjective variation, the overall consistency remains comparable to human agreement. Therefore, averaging multiple LLM judges provides a reasonable and scalable approximation of human diagnostic evaluation for Task 2.

\section{Case Study}
\begin{figure*}[t]
\centering
\resizebox{\linewidth}{!}{%
  \includegraphics{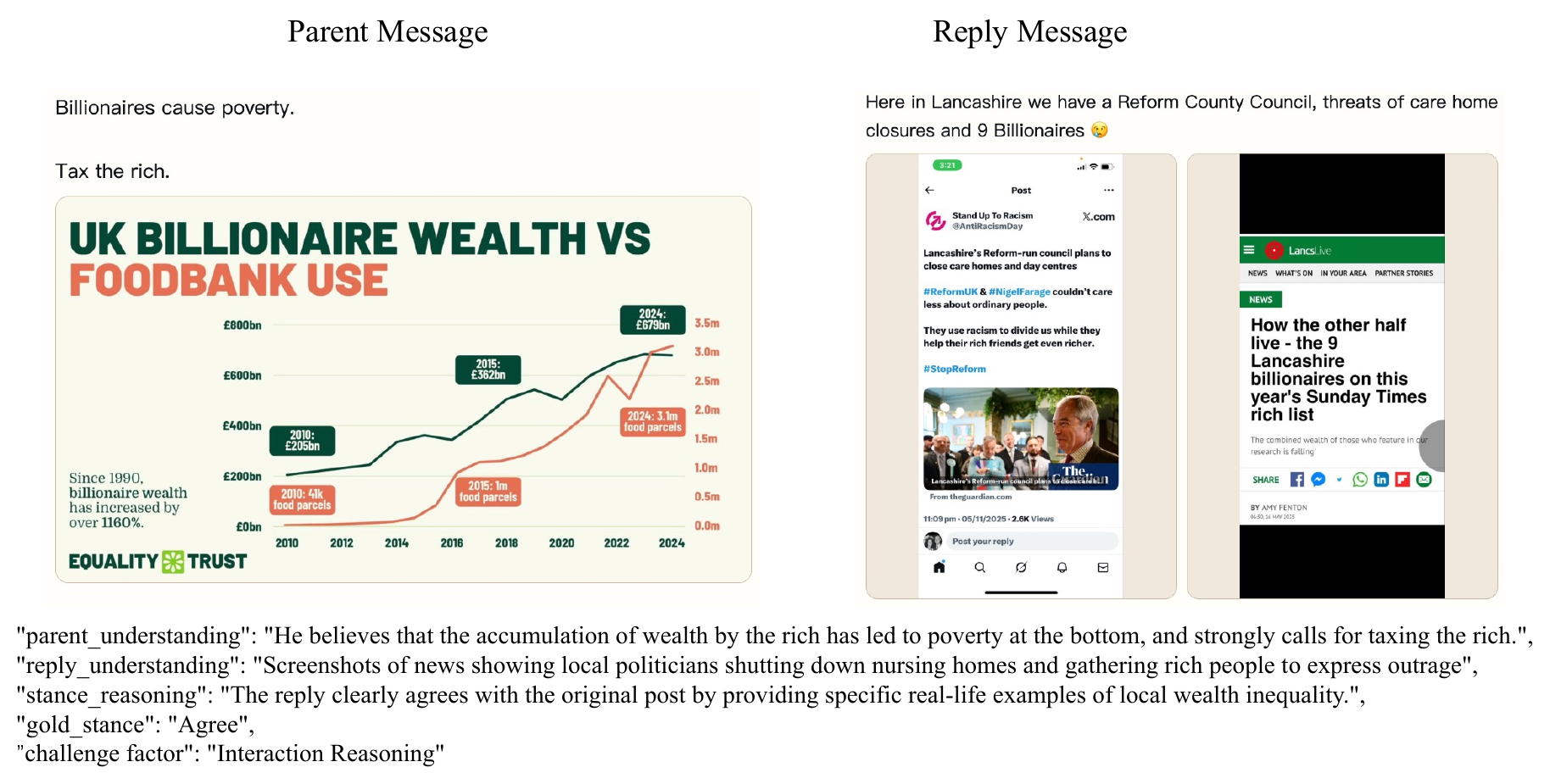}
}
\caption{Case study.}
\label{fig:case1}
\end{figure*}
To provide a concrete example of MMDS-Bench, Figure~\ref{fig:case1} shows a representative multimodal parent--reply case with its gold dynamic stance label, annotated challenge factors, and reference diagnostic reasoning.


\begin{table*}[t]
\centering
\small
\resizebox{0.75\textwidth}{!}{
\begin{tabular}{lcccccccc}
\toprule
\multirow{2}{*}{\textbf{Score}} 
& \multicolumn{4}{c}{\textbf{Human Annotators}} 
& \multicolumn{4}{c}{\textbf{LLM Judges}} \\
\cmidrule(lr){2-5}
\cmidrule(lr){6-9}
& \textbf{H1--H2} 
& \textbf{H1--H3} 
& \textbf{H2--H3} 
& \textbf{Avg.} 
& \textbf{Gi--Qw} 
& \textbf{Gi--Ge} 
& \textbf{Qw--Ge} 
& \textbf{Avg.} \\
\midrule
PUS & 0.57 & 0.73 & 0.69 & 0.66 & 0.56 & 0.77 & 0.65 & 0.66 \\
RUS & 0.70 & 0.71 & 0.73 & 0.71 & 0.56 & 0.83 & 0.61 & 0.67 \\
SRS & 0.87 & 0.90 & 0.87 & 0.88 & 0.90 & 0.94 & 0.91 & 0.92 \\
\bottomrule
\end{tabular}
}
\caption{
Pairwise agreement among human annotators and LLM judges on the human validation subset. Agreement is measured using Spearman correlation over 600 prediction cases. Gi, Qw, and Ge denote Gemini, Qwen, and Gemma, respectively. PUS, RUS, and SRS denote Parent Understanding Score, Reply Understanding Score, and Stance Reasoning Score.
}
\label{tab:app_pairwise_agreement_human_llm}
\end{table*}


\begin{table}[t]
\centering
\small
\resizebox{0.9\linewidth}{!}{
\begin{tabular}{llccc}
\toprule
\multirow{2}{*}{\textbf{Score}} 
& \multirow{2}{*}{\textbf{Human}} 
& \multicolumn{3}{c}{\textbf{LLM Judge}} \\
\cmidrule(lr){3-5}
& 
& \textbf{Gemini} 
& \textbf{Qwen} 
& \textbf{Gemma} \\
\midrule
\multirow{3}{*}{PUS}
& H1 & 0.465 & 0.667 & 0.541 \\
& H2 & 0.615 & 0.620 & 0.640 \\
& H3 & 0.526 & 0.729 & 0.614 \\
\midrule
\multirow{3}{*}{RUS}
& H1 & 0.640 & 0.515 & 0.715 \\
& H2 & 0.720 & 0.566 & 0.780 \\
& H3 & 0.630 & 0.635 & 0.707 \\
\midrule
\multirow{3}{*}{SRS}
& H1 & 0.905 & 0.891 & 0.926 \\
& H2 & 0.848 & 0.837 & 0.850 \\
& H3 & 0.873 & 0.882 & 0.878 \\
\bottomrule
\end{tabular}
}
\caption{
Detailed human--LLM judge agreement on the human validation subset. Each cell under ``LLM Judge'' reports the Spearman correlation between one human annotator and one LLM judge over 600 prediction cases. PUS, RUS, and SRS denote Parent Understanding Score, Reply Understanding Score, and Stance Reasoning Score, respectively.
}
\label{tab:app_human_llm_cross_agreement}
\end{table}

\end{document}